\documentclass{article}

\PassOptionsToPackage{numbers, compress}{natbib}

\usepackage[preprint]{neurips_2023}

\usepackage[utf8]{inputenc} 
\usepackage[T1]{fontenc}    
\usepackage{hyperref}       
\usepackage{url}            
\usepackage{booktabs}       
\usepackage{amsfonts}       
\usepackage{nicefrac}       
\usepackage{microtype}      
\usepackage{xcolor}         

\usepackage{bm}
\usepackage{dsfont}
\usepackage{colortbl}
\usepackage{multirow}
\usepackage{graphicx}
\usepackage{verbatim}
\usepackage{amsmath}
\usepackage{subfigure}
\usepackage{amssymb}
\definecolor{mygray}{gray}{.9}
\newcommand{\etal}{\textit{et al}.}
\newcommand{\ie}{\textit{i}.\textit{e}.}
\newcommand{\eg}{\textit{e}.\textit{g}.}

\title{OpenGCD: Assisting Open World Recognition with Generalized Category Discovery}

\author{
	Fulin Gao, Weimin Zhong\thanks{Corresponding author}\:\:, Zhixing Cao, Xin Peng$^{\ast}\textbf{,}$ Zhi Li\\
	Key Laboratory of Smart Manufacturing in Energy Chemical Process, Ministry of Education\\
	East China University of Science and Technology\\
	Shanghai 200237, China\\
	\texttt{fulingao@mail.ecust.edu.cn, \{wmzhong, zcao,  xinpeng, zhili\}@ecust.edu.cn}\\
}

\begin{document}	
	\maketitle
	
	\begin{abstract}
		A desirable open world recognition (OWR) system requires performing three tasks: (1) Open set recognition (OSR), \ie, classifying the known (classes seen during training) and rejecting the unknown (unseen$/$novel classes) \emph{online}; (2) Grouping and labeling these unknown as novel known classes; (3) Incremental learning (IL), \ie, continual learning these novel classes and retaining the memory of old classes. Ideally, all of these steps should be automated. However, existing methods mostly assume that the second task is completely done manually. To bridge this gap, we propose OpenGCD that combines three key ideas to solve the above problems sequentially: (a) We score the origin of instances (unknown or specifically known) based on the uncertainty of the classifier's prediction; (b) For the first time, we introduce generalized category discovery (GCD) techniques in OWR to assist humans in grouping unlabeled data; (c) For the smooth execution of IL and GCD, we retain an equal number of informative exemplars for each class with diversity as the goal. Moreover, we present a new performance evaluation metric for GCD called harmonic clustering accuracy. Experiments on two standard classification benchmarks and a challenging dataset demonstrate that OpenGCD not only offers excellent compatibility but also substantially outperforms other baselines. Code: \url{https://anonymous.4open.science/r/OpenGCD-61F6/}.
	\end{abstract}

	\section{Introduction}\label{sec:intro}
	Human cognition is the process of transforming, storing, learning and using the information received continually. For example, a child born in Asia will naturally recognize pandas, elephants and rhinoceroses. If he arrives in Australia, although he cannot recognize kangaroos and koalas, he can still identify them as unseen and two different animals based on his prior knowledge and their characteristics. After learning from his parents or others, he knows what species both are. In order not to forget these animals, he also takes pictures. In this way, this child can find and distinguish unseen animals according to his prior knowledge and their characteristics, and later recognize them by learning, and permanently remember these seen animals through photos. Inspired by this, several recent studies \cite{Bendale-15,Rudd,Xu,Mancini,Fontanel} have attempted to theorize this human mind and formulated an architecture called open world recognition (OWR).
	
	A desirable OWR system requires performing three main tasks: (1) Open set recognition (OSR), \ie, classifying the known (classes seen during training) and rejecting the unknown (unseen$/$novel classes) \emph{online}; (2) Grouping and labeling these unknown as novel known classes; (3) Incremental learning (IL), \ie, continual learning these novel classes and retaining the memory of old classes \cite{Bendale-15,Rudd}. In this paper, we propose an approach called \emph{assisting {\bf\emph{open}} world recognition with {\bf\emph{g}}eneralized {\bf\emph{c}}ategory {\bf\emph{d}}iscovery} (OpenGCD) that combines three key ideas to address the above tasks sequentially. 
	
	For the first task, \ie, OSR, thresholding the closed set predictions of a classifier and evaluating the likelihood that a instance is from an unknown class based on the marginal distribution are two popular options \cite{Geng}. The former is lightweight, while the latter is intuitive. Inspired by this, our first idea is to develop an OSR method that combines the advantages of both. To this end, we evaluate the likelihood that a instance is from an unknown class based on the uncertainty of the classifier's closed set prediction. It is not only computationally lightweight as thresholding methods, but also allows to visualize the probability distribution of the instance over unknown and all known classes as evaluation methods.
	
	For the second task, since Bendale and Boult \cite{Bendale-15} first formalized the OWR problem, the vast majority of subsequent work has followed their setting to solve this task exclusively manually, \eg, \cite{Rudd,Xu,Mancini,Fontanel}. It is laborious and expensive. Furthermore, we find that it is essentially a task to classify all data in the unlabeled set (rejected instance set) given a labeled dataset (available training set). Ideally, labeled and unlabeled datasets are class-disjoint. At this point, this task coincides with novel category discovery (NCD), with the difference that the latter only requires clustering unlabeled data, while the former requires further specifying explicit classes. However, the fact is that there are always some instances from known classes that are falsely rejected, \ie, labeled and unlabeled datasets may be class-intersecting. As an extension to NCD, generalized category discovery (GCD) takes this into account. Inspired by this, our second idea is to introduce GCD techniques to assist humans in grouping unlabeled data. To this end, we employ the semi-supervised $k$-means++ (ss-$k$-means++) algorithm \cite{Jordan} to filter and group instances from novel classes in the unlabeled dataset. Thus, the labeler simply picks out the obviously incompatible instances from each group, rather than struggling to label the messy data directly.

	However, this approach requires knowledge of the total number of classes for both known and novel classes, which is not realistic in the open world. To this end, we fine-tune the class number estimation protocol proposed by Han \etal \cite{Han-A} to allow it to accelerate the search process by Brent's algorithm as in \cite{Vaze}. Moreover, we find that the average clustering accuracy (ACC) \cite{Han-L}, an evaluation metric still widely used in NCD and GCD until now \cite{Han-A,Vaze,Zhang,Zhao,Cao}, fails to distinguish explicitly between known and novel classes, resulting in improper evaluation. Thus, we extend ACC to the harmonic clustering accuracy (HCA), which measures known and novel classes with classification accuracy and ACC, respectively, and then harmonizes the two.
	
	For the third task, \ie, IL, the challenge lies in acquiring novel knowledge while avoiding catastrophic forgetting of old knowledge. After all, in an open dynamic world, full training data may only be temporarily available due to storage constraints or privacy concerns \cite{DeLange-A}. Furthermore, GCD cannot function smoothly without informative labeled data from known classes. Fortunately, we find the popular replay technique to be a straightforward yet effective solution. Inspired by this, our third idea is to select some informative exemplars when data are available and save them for subsequent GCD and IL. To this end, we employ the dissimilarity-based sparse subset selection (DS3) algorithm \cite{Elhamifar} for exemplar selection to ensure the diversity of pre-stored instances. Compared to methods that aim at selecting exemplars for representativeness, \eg, \cite{Zhang,Liu}, it preserves as much spatial information as possible from the original data thus reducing open space risk, \ie, the risk of classifying known instances into unknown.
	
	Overall, the contributions of this work can be highlighted as follows: (\romannumeral1) A highly compatible OWR scheme dubbed OpenGCD is provided, which is independent of classifier, so any well-designed closed set classifier can be easily embedded in it for OWR; (\romannumeral2) GCD is first introduced to assist the task of filtering and grouping unlabeled data in OWR to reduce labor costs, which drives OWR another small step towards automation; (\romannumeral3) A new performance evaluation metric called HCA is presented for NCD and GCD, which solves the problem that ACC fails to distinguish explicitly between known and novel classes resulting in improper evaluation; (\romannumeral4) A thorough empirical evaluation of OpenGCD is reported, showing significant performance improvements in various tasks of OWR.
	
	\section{Related work}\label{sec:rela}
\begin{figure}[!htbp]
		\centering
		\includegraphics[width=0.7\linewidth]{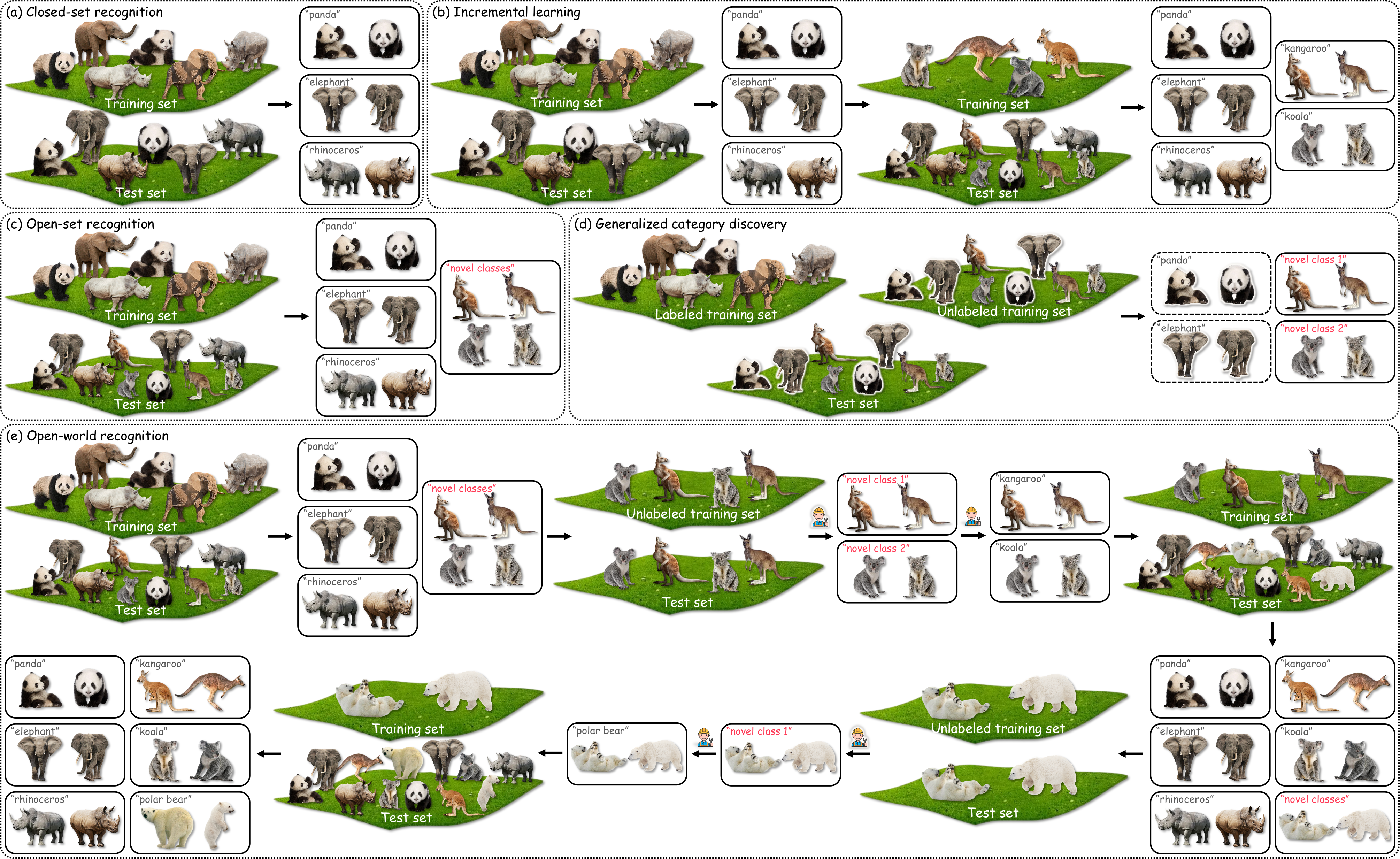}
		\caption{Schematic of various problems in the open world.}
		\label{fig:methods}
	\end{figure}

	We visually illustrate the similarities and differences between OWR and related settings in Fig. \ref{fig:methods}. Next, we briefly review the most representative related works.
	
	\vspace{-4mm}
	\paragraph{Open set recognition} 
	As shown in Fig. \ref{fig:methods}(c), in the OSR scenario, incomplete knowledge of the world exists in the training set and unknown classes can be submitted to the system during testing. It requires the \emph{online} model not only to \emph{classify} the known$/$seen classes, but also to \emph{reject} the unknown$/$unseen$/$novel ones \cite{Wang,Yang}. 1-vs-all principle, thresholding, and unknown probability estimation are the three most popular OSR strategies \cite{Geng}. 1-vs-all principle-based methods \cite{Scheirer,Mendes-O} are the earliest in origin but relatively cumbersome. The threshold-based methods  \cite{Jain,Mendes-N} offer high compatibility and low computational overhead. The methods \cite{Bendale-16,Gao} for estimating unknown probability are the most intuitive. 
	
	\vspace{-4mm}
	\paragraph{Generalized category discover}
	As shown in Fig. \ref{fig:methods}(d), in the GCD scenario, the unlabeled test set (available for training) may contain both classes that have been seen and unseen during training. It requires the model not only to \emph{classify} known$/$seen classes, but also to \emph{cluster} unknown$/$unseen$/$novel ones \cite{Zhao}. Its three major differences from OSR are whether it supports online runs, whether the unlabeled test set is available for training, and whether the unknown classes should be rejected or clustered. Furthermore, if the test and training sets are class-disjoint, the problem degenerates into NCD, which can be illustrated by Fig. \ref{fig:methods}(d) with the white-emitting animals removed. As an emerging technology, representative works are \cite{Han-A,Vaze,Zhang,Cao}.
	
	\vspace{-4mm}
	\paragraph{Incremental learning}  
	As shown in Fig. \ref{fig:methods}(b), in the IL scenario, instead of unseen classes, novel known classes are submitted to the system during testing. Moreover, the full training data from old known classes may only be temporarily available due to storage constraints or privacy concerns. It requires the model to continuously learn knowledge of novel known classes while avoiding catastrophic forgetting of old known classes \cite{Sun}. Regularization, parameter isolation and replay are the three most popular techniques \cite{DeLange-A}. The former two \cite{Li,Mallya} offer low compatibility due to their strong dependence on neural network classifiers. The last one \cite{DeLange-C,Liu} is simple but effective and essential for GCD.
	
	\vspace{-4mm}
	\paragraph{Open world recognition}
	As shown in Fig. \ref{fig:methods}(e), in the OWR scenario, the settings of OSR and IL are perfectly followed. The GCD setting will also be catered if the replay IL scheme is adopted. It requires the model to OSR, group and label unlabeled data, and IL in sequence. As a challenging task, representative works are \cite{Bendale-15,Rudd,Xu,Mancini,Fontanel}. Interestingly, they both adopted thresholding methods for OSR and processed unlabeled data manually. Inspired by this, we developed OpenGCD, whose flow can be illustrated by Fig. \ref{fig:methods}(e), in which white-emitting workers are replaced by GCD.

	\section{Assisting open world recognition with generalized category discovery}\label{sec:OpenGCD}

	\paragraph{Problem Formulation} A solution to OWR is a tuple $[\mathcal{O},\mathcal{L},\mathcal{I}]$ with:
	\begin{enumerate}\setlength{\itemsep}{0pt}
		\item[1)] An OSR function $\mathcal{O}\!:\!\mathbb{R}^{3\times H\times W}\!\mapsto\!\mathbb{N}$. Given an \emph{online} instance set $\mathcal{X}\!\subset\!\mathbb{R}^{3\times H\times W}$, $\mathcal{O}$ should assign $\boldsymbol{x}\!\in\!\mathcal{X}$ to either $\mathcal{C}_t^l\!\subset\!\mathbb{N}^+$ (known classes at phase $t$) or $0$ (unknown classes). See Secs. \ref{sec:fe}-\ref{sec:OSR} for our $\mathcal{O}$.
		\item[2)] A labeling process $\mathcal{L}\!:\!\mathbb{R}^{3\times H\times W}\!\mapsto\!\mathbb{N}^+$. Given an unlabeled unknown instance set $\mathcal{X}^0\!\subset\!\mathcal{X}$, $\mathcal{L}$ should assign ground-truth labels to $\boldsymbol{x}^0\!\in\!\mathcal{X}^0$. Assuming that the novel classes discovered are $\mathcal{C}_t^n\!\subset\!\mathbb{N}^+$ where $\mathcal{C}_t^n\cap\mathcal{C}_t^l=\varnothing$, then it yields $\mathcal{C}_{t+1}^l=\mathcal{C}_t^l\cup\mathcal{C}_t^n$. See Sec. \ref{sec:magcd} for our $\mathcal{L}$.
		\item[3)] An IL function $\mathcal{I}_t:\mathcal{H}^{|\mathcal{C}_t^l|}\!\mapsto\!\mathcal{H}^{|\mathcal{C}_{t+1}^l|}$. Given a labeled instance set $\mathcal{X}^n\!\subset\!\mathcal{X}^0$ of novel classes, $\mathcal{I}_t$ should allow $\mathcal{O}$ to learn $\mathcal{C}_t^n$ and retain the ability to recognize $\mathcal{C}_t^l$. See Sec. \ref{sec:il} for our $\mathcal{I}_t$.
	\end{enumerate}

	\subsection{Feature embedding}\label{sec:fe}
	Given an instance $\boldsymbol{x}\!\in\!\mathbb{R}^{3\times H\times W}$, the goal of feature embedding is to convert it into a flat feature $\boldsymbol{z}\!\in\!\mathbb{R}^D$. The benefit is that it gives an interface allowing us to design subsequent models as we wish. It is possible to add the classification head or plug in any other type of classifier, \eg, support vector machine (SVM), XGBoost.
	
	The features generated by the vision transformer (ViT) \cite{Caron} with self-supervised contrastive learning offer discriminative spatial representations. Thus, as in \cite{Vaze}, we employ ViT trained on the \emph{unlabeled} ImageNet with DINO self-supervision as the feature extractor $f\!:\!\mathbb{R}^{3\times H\times W}\!\mapsto\!\mathbb{R}^D$. We can get the feature representation $\boldsymbol{z}$ of the instance $\boldsymbol{x}$ via $f$. All our subsequent procedures are executed on features extracted from the frozen ViT.
	
	\subsection{Exemplar selection}\label{sec:es}
	At phase $t$, given a temporarily available labeled feature set $\mathcal{Z}_t^l\!=\!\{\boldsymbol{z}^l_i,y^l_i\}_{i=1}^{N_t}$ where $y^l_i\!\in\!\mathcal{C}_t^l\!\subset\!\mathbb{N}^+$, the goal of the exemplar selection is to retain informative instances of it for GCD (Sec. \ref{sec:gcd}) and IL (Sec. \ref{sec:il}).
	
	We apply the DS3 algorithm \cite{Elhamifar} to select exemplars $\mathcal{E}_t$ from $\mathcal{Z}_t^l$ and store them in buffer $\mathcal{M}_r$. DS3 defines the objective function based on the difference between instances and solves it by the alternating direction method of multipliers (ADMM). DS3 is only our default choice because of its ability to preserve diverse and informative instances. The exemplars selected with the goal of diversity retained as much spatial information as possible from the original data, avoiding the expansion of unknown spaces and thus reducing the open space risk, \ie, the risk of categorizing known instances as unknown. In fact, any similar exemplar selection approach is an alternative. Moreover, to avoid out of memory, $|\mathcal{M}_r|$ is always fixed at $N_0$, the size of the memory occupied by the data in the initial phase. To ensure class balance, DS3 is executed once on the feature subset of each known class, so that $|\mathcal{M}_r|/|\mathcal{C}_t^l|$ exemplars are retained for each known class.
	
	\subsection{Classifier (re)fitting}\label{sec:fit}
	Given the labeled exemplar set $\mathcal{E}_t\!=\!\{\boldsymbol{z}^e_i,y^e_i\}_{i=1}^{N_0}\!\subseteq\!\mathcal{Z}_t^l$ where $y^e_i\!\in\!\mathcal{C}_t^l\!\subset\!\mathbb{N}^+$, the goal of (re)fitting the classifier is to allow the classifier to learn the existing knowledge of the known classes. Regardless of the current phase, this is a process from scratch. It is uncomplicated benefiting from the fact that the total number of exemplars is constant and features are (re)fitted directly instead of instances. 
	
	We choose an appropriate classifier $\varphi\!:\!\mathbb{R}^{D}\!\mapsto\!\mathbb{R}^{|\mathcal{C}_t^l|}$ to fit on $\mathcal{E}_t$. Since OpenGCD has no dependency on classifier type, any well-designed classifier is an alternative. Considering that this is not the focus of this study, we take the multilayer perceptron (MLP, which can be considered as a classification head) or the SVM and XGBoost with default parameters as candidates.
	
	\subsection{Uncertainty-based open set recognition}\label{sec:OSR}	
	Given an unlabeled feature set $\mathcal{Z}_t^u\!=\!\{\boldsymbol{z}^u_i\}_{i=1}^{M_t}$ (allowing continuous online delivery), the goal of OSR is to assign $label$ {\em0} to features from unknown classes and assign other features to $\mathcal{C}_t^l$. 
	
	By feeding $\mathcal{Z}_t^u$ into $\varphi$, the predicted probability distribution $\mathcal{P}_t\!=\!\{\boldsymbol{p}_i\}_{i=1}^{M_t}$ over each known classes is obtained. Let the label set of $\mathcal{Z}_t^u$ be $\mathcal{C}_t^u\!\subset\!\mathbb{N}^+$. If $\mathcal{C}_t^u\!\subseteq\!\mathcal{C}_t^l$ is known, the goal degenerates to closed set recognition (CSR) and the predicted labels can be assigned by $y_i^\ast\!=\!{\arg\max}_{j\in\mathcal{C}_t^l}p_{ij}$; otherwise, OSR should be initiated.
	
	The less confident a classifier is in predicting a feature, the higher the likelihood that the instance corresponding to this feature is from an unknown class. We propose to capture this diffidence through uncertainty. Thus, we approximate the uncertainty of the classifier's prediction $\boldsymbol{p}_i$ by:\vspace{-1mm}
	\begin{equation}\label{eq:u}
		u_i=1-\max_{j\in\mathcal{C}_t^l} p_{ij}
	\end{equation}
	
	Then, we define the unknown probability as:\vspace{-1mm}
	\begin{equation}\label{eq:p0}
		p_{i0}=\alpha\times u_i
	\end{equation}
	where $\alpha$ is a regulatory factor to control the temperature of the uncertainty. 
	
	Next, we can get a new probability distribution by:\vspace{-1mm}
	\begin{equation}\label{eq:p_hat}
		p^\prime_{ij}=\frac{e^{p_{ij}}}{\sum_{j=0}^{|\mathcal{C}_t^l|}e^{p_{ij}}}
	\end{equation}
	
	So far, we have visualized the probability distribution of the instance over the unknown ($0$) and all known classes ($\mathcal{C}_t^l$). Moreover, it is clear from Eqs. \eqref{eq:u}-\eqref{eq:p_hat} that the related operations are quite lightweight, so the computational overhead is low. Essentially, it is along the same lines as the approach of thresholding $\boldsymbol{p}_i$, which is to reject instances with low maximum prediction probability, while our approach is more intuitive. Specifically, in the case of $\boldsymbol{p}_i=[0.1,0.2,0.6]$, the thresholding approach can reject $\boldsymbol{z}^u_i$ by setting the threshold to be greater than $0.6$ without quantitatively characterizing the likelihood that $\boldsymbol{z}^u_i$ falling into the unknown. Whereas our approach can describe this likelihood by $p^\prime_{i0}$ and reject $\boldsymbol{z}^u_i$ based on $p^\prime_{i0}>p^\prime_{ij},j\in\mathcal{C}_t^l$.
	
	Finally, the predicted labels can be assigned by $y_i^\ast\!=\!{\arg\max}_{j\in\{0,\mathcal{C}_t^l\}}p^\prime_{ij}$. Let the feature subset in $\mathcal{Z}_t^u$ with predicted label of $0$ be $\mathcal{Z}_t^0$. 
	
	\subsection{Assisting manual annotation with generalized category discovery}\label{sec:magcd}	
	Given the labeled exemplar set $\mathcal{E}_t\!=\!\{\boldsymbol{z}^e_i,y^e_i\}_{i=1}^{N_0}\!\subseteq\!\mathcal{Z}_t^l$ where $y^e_i\!\in\!\mathcal{C}_t^l\!\subset\!\mathbb{N}^+$ and the rejected unlabeled feature set $\mathcal{Z}_t^0\!=\!\{\boldsymbol{z}^0_i\}_{i=1}^{M^0_t}\!\subseteq\!\mathcal{Z}_t^u$ (available for training), the goal of assisting manual annotation with GCD is to first automatically filter and group features from novel classes in $\mathcal{Z}_t^0$ using GCD, followed by manual correction and labeling.
	
	\subsubsection{Generalized category discovery}\label{sec:gcd}
	As with the input of Sec. \ref{sec:magcd}, the goal of GCD is to put the falsely rejected features in $\mathcal{Z}_t^0$ back to $\mathcal{C}_t^l$ and cluster other features in $\mathcal{Z}_t^0$. We introduce GCD to assist in manually filtering and grouping features.
	
	Let the novel classes in $\mathcal{Z}_t^0$ be $\mathcal{C}_t^n\!\subset\!\mathbb{N}^+$, and all classes in $\mathcal{E}_t\!\cup\!\mathcal{Z}_t^0$ be $\mathcal{C}_t\!=\!\mathcal{C}_t^l\cup\mathcal{C}_t^n$ where $\mathcal{C}_t^l\cap\mathcal{C}_t^n\!=\!\varnothing$. However, we usually have no prior knowledge of $|\mathcal{C}_t^n|$ or $|\mathcal{C}_t|$. Here, we use the estimated $|\hat{\mathcal{C}}_t|$ instead of $|\mathcal{C}_t|$ (the estimation problem is solved in Sec. \ref{sec:k}). Afterwards, we employ ss-$k$-means++ \cite{Jordan} with $\mathcal{E}_t$ as supervision to filter and group features from novel classes in $\mathcal{Z}_t^0$ as in \cite{Vaze}. ss-$k$-means++ determines the centroids of $|\mathcal{C}_t^l|$ known classes by $\mathcal{E}_t$ and selects the remaining $|\hat{\mathcal{C}}^n_t|$ ($|\hat{\mathcal{C}}^n_t|\!=\!|\hat{\mathcal{C}}_t|\!-\!|\mathcal{C}_t^l|$) centroids with a probability proportional to the distance from the feature to the nearest centroid. At each iteration, we force the data in $\mathcal{E}_t$ to map to the ground-truth labels. Likewise, ss-$k$-means++ is only our default choice, and any approach with semi-supervised clustering capabilities is an option. 
	
	Finally, let the labels assigned to $\mathcal{Z}_t^0$ by ss-$k$-means++ be $\{\hat{y}^0_i\}_{i=1}^{M^0_t}$ where $\hat{y}^0_i\in\mathcal{C}_t^l\cup\mathcal{N}_t$ and $\mathcal{N}_t$ are $|\hat{\mathcal{C}}^n_t|$ novel groups that are clustered. Then, let the features from $\mathcal{N}_t$ in $\mathcal{Z}_t^0$ and the corresponding predicted labels be combined into $\hat{\mathcal{Z}}_t^n$.
	
	\subsubsection{Manual annotation}\label{sec:ma}
	Given the feature set $\hat{\mathcal{Z}}_t^n\!=\!\{\boldsymbol{z}^n_i,\hat{y}^n_i\}_{i=1}^{\hat{M}^n_t}$ with predicted cluster labels where $\boldsymbol{z}^n_i\!\in\!\mathcal{Z}_t^0$ and $\hat{y}^n_i\!\in\!\mathcal{N}_t$, the goal of manual annotation is to correct and label each cluster.
	
	Engineers can fetch the instances corresponding to $\hat{\mathcal{Z}}_t^n$, then visually locate the distinctive images in each cluster without much effort and put them into other appropriate clusters, and finally assign ground-truth labels. Of course, it is also necessary to remove features from $\mathcal{C}_t^l$ in $\hat{\mathcal{Z}}_t^n$ since there is no perfect model. Moreover, if $|\mathcal{C}^n_t|\!>\!|\hat{\mathcal{C}}^n_t|$ or $|\mathcal{C}^n_t|\!<\!|\hat{\mathcal{C}}^n_t|$, the corresponding clusters need to be added or removed. Nevertheless, compared with processing instances one by one, filtering and grouping data by GCD technology can still significantly reduce labor costs.
	
	Let the combination of the features in $\hat{\mathcal{Z}}_t^n$ from novel classes $\mathcal{C}^n_t$ and the corresponding ground-truth labels be $\mathcal{Z}_t^n$.
	
	\subsubsection{Estimating the number of classes}\label{sec:k}
	As with the input of Sec. \ref{sec:magcd}, the goal of estimating the number of classes is to determine $k$ in ss-$k$-means++, \ie, $|\mathcal{C}_t|$.
	
	We observe that the class number estimation protocol in \cite{Vaze} improves the search efficiency by Brent's algorithm but is prone to fall into the greedy trap by using ACC as the only evaluation metric. Conversely, \cite{Han-A} circumvents this problem by evaluating labeled and unlabeled predictions separately but executes inefficiently due to traversal search. Thus, we fine-tune the protocol in \cite{Han-A} to allow it to accelerate the search process by Brent's algorithm as in \cite{Vaze}. 
	
	Specifically, we first split $\mathcal{E}_t$ into an anchor set $\mathcal{E}^a_t$ with classes $\mathcal{C}^a_t$ and a validation set $\mathcal{E}^v_t$ with classes $\mathcal{C}^v_t$ where $\mathcal{C}^a_t\cup \mathcal{C}^v_t\!=\!\mathcal{C}^l_t$, $\mathcal{C}^a_t\cap \mathcal{C}^v_t\!=\!\varnothing$, $|\mathcal{C}^a_t|\!:\!|\mathcal{C}^v_t|\!=\!2\!:\!1$. Then, we launch Brent's algorithm to execute ss-$k$-means on $\mathcal{C}^a_t\cup\mathcal{C}^v_t\cup\mathcal{Z}_t^0$ bounded by $(|\mathcal{C}^a_t|,|\mathcal{C}^{\max}_t|]$ until convergence. $|\mathcal{C}^{\max}_t|$ is an expected maximum number of total classes, and it is allowed to set a large value if there is no this knowledge. During semi-supervised learning, features in $\mathcal{E}^a_t$ are forced to follow ground-truth labels and features in $\mathcal{E}^v_t$ are considered as additional ``unlabeled" data. The clustering performance of $\mathcal{E}^v_t$ and $\mathcal{Z}_t^0$ is evaluated using ACC and silhouette coefficient (SC), given below, respectively, and Brent's algorithm takes maximizing ACC+SC as the optimization objective. Finally, Brent's algorithm terminates at the optimal estimate $|\hat{\mathcal{C}}_t|$. 
	
	The two main differences between our protocol and the original one of \cite{Han-A} are whether the centroids of novel classes are initialized by $k$-means++ and whether the search process is accelerated by Brent's algorithm. 
	
	\vspace{-4mm}
	\paragraph{Cluster quality indices}
	The first index is ACC, which is applicable to the $\mathcal{C}_t^v$ labeled classes in the validation set $\mathcal{E}^v_t$ and is given by:\vspace{-1mm}
	\begin{equation}\label{eq:ACC}
		\text{ACC}=\max_{g\in\mathcal{G}(\mathcal{C}_t^v)}\frac{1}{L}\sum_{i=1}^{L} \mathds{1}\{y^v_i=g(\hat{y}_i^v)\}
	\end{equation}
	where $y_i^v$ and $\hat{y}_i^v$ denote the ground-truth label and clustering assignment for each feature $\boldsymbol{z}_i^v$ in $\mathcal{E}^v_t$, $L\!=\!|\mathcal{E}^v_t|$, and $\mathcal{G}(\mathcal{C}_t^v)$ is the group of permutations of $|\mathcal{C}_t^v|$ elements (this discounts the fact that the cluster indices may not be in the same order as the ground-truth labels). Permutations are optimized using the Hungarian algorithm \cite{Harold}.
	
	The other index is SC, which is applicable to the unlabeled features $\mathcal{Z}_t^0$ and is given by:\vspace{-1mm}
	\begin{equation}\label{eq:SC}
		\text{SC}=\sum_{i=1}^{M_t^0}\frac{b(\boldsymbol{z}_i^0)-a(\boldsymbol{z}_i^0)}{\max\{a(\boldsymbol{z}_i^0),b(\boldsymbol{z}_i^0)\}}
	\end{equation}
	where $a(\boldsymbol{z}_i^0)$ is the average distance between $\boldsymbol{z}_i^0$ and all other features within the same cluster, and $b(\boldsymbol{z}_i^0)$ is the smallest average distance of $\boldsymbol{z}_i^0$ to all features in any other cluster (of which $\boldsymbol{z}_i^0$ is not a member).
	
	\subsection{Exemplar-based incremental learning}\label{sec:il}
	Given the labeled novel class feature set $\mathcal{Z}_t^n\!=\!\{\boldsymbol{z}^n_i,y^n_i\}_{i=1}^{M^n_t}$ where $\boldsymbol{z}^n_i\!\in\!\hat{\mathcal{Z}}_t^n$ and $y^n_i\!\in\!\mathcal{C}_t^n\!\subset\!\mathbb{N}^+$ and the labeled old class exemplar set $\mathcal{E}_t\!=\!\{\boldsymbol{z}^e_i,y^e_i\}_{i=1}^{N_0}\!\subseteq\!\mathcal{Z}_t^l$ where $y^e_i\!\in\!\mathcal{C}_t^l\!\subset\!\mathbb{N}^+$, the goal of IL is for the classifier to continuously learn knowledge of novel classes $\mathcal{C}_t^n$ and retain memory of old classes $\mathcal{C}_t^l$.
	
	Thus, we merge $\mathcal{E}_t$ and $\mathcal{Z}_t^n$ to get the labeled feature set $\mathcal{Z}_{t+1}^l\!=\!\{\boldsymbol{z}^l_i,y^l_i\}_{i=1}^{N_{t+1}}$ where $N_{t+1}\!=\!N_0\!+\!M^n_t$ and $y^l_i\in\mathcal{C}_{t+1}^l\!=\!\mathcal{C}_t^l\cup\mathcal{C}_t^n$, for stage $t\!+\!1$. Since $\mathcal{Z}_{t+1}^l$ is also temporarily available and the number of features of the novel classes may differ significantly from that of the old classes, exemplar selection is required. Before that, we let $t\!=\!t\!+\!1$ to enter Sec. \ref{sec:es} to formally launch the next stage.
	
	The schematic of the formulated OpenGCD is shown in Appendix A. 

	\section{Experiments}
	\subsection{Experimental setup}
	\subsubsection{Data}
	We evaluate OpenGCD on two standard benchmark datasets CIFAR10 \cite{Alex}, CIFAR100 \cite{Alex}, and a challenging dataset CUB \cite{Wah}. CIFAR10$/$CIFAR100$/$CUB contains 50,000$/$50,000$/$5,994 training images and 10,000$/$10,000$/$5,794 test images from 10$/$100$/$200 classes. Since ViT is self-supervised trained on ImageNet \cite{Deng}, we do not involve ImageNet in our test experiments as it is not completely unknown to OpenGCD.

	\subsubsection{Metrics}
	We adopt accuracy, harmonic normalized accuracy (HNA) \cite{Mendes-O}, and HCA to evaluate the performance of IL, OSR, and GCD, respectively. 
	
	\vspace{-4mm}
	\paragraph{Accuracy}
	A widely used metric for evaluating CSR performance, given by:\vspace{-1mm}
	\begin{equation} \label{eq:Acc}  
		\text{Acc}=\frac{1}{M}\sum_{i=1}^{M}\mathds{1}\{y_i=\hat{y}_i\}
	\end{equation}
	where $M$ is the number of test instances, $y_i$ and $\hat{y}_i$ are the ground-truth and predicted labels. We adopt it to show the performance degradation of the same test set at different phases. The more severe the degradation, the more the IL fails.
	
	\vspace{-4mm}
	\paragraph{Harmonic normalized accuracy}
	A widely used metric for evaluating OSR performance, given by:\vspace{-1mm}
	\begin{equation}\label{eq:HAC}\begin{split}
			\text{HNA}=\begin{cases}0,&if~\text{AKS}=0~or~\text{AUS}=0\\\frac{2}{\left(\frac{1}{\text{AKS}}+\frac{1}{\text{AUS}}\right)}, &otherwise\end{cases}
	\end{split}\end{equation} 
	where AKS and AUS are the accuracy of known and unknown classes calculated by Eq. \ref{eq:Acc}, respectively. It harmonizes AKS and AUS, and its higher score indicates more successful OSR.
	
	\vspace{-4mm}
	\paragraph{Harmonic clustering accuracy}
	A new metric for evaluating NCD or GCD performance, yielded by extending ACC, given by:\vspace{-1mm}
	\begin{equation}\label{eq:HAC}\begin{split}
			\text{HCA}=\begin{cases}0,&if~\text{AKS}=0~or~\text{ANS}=0\\\frac{2}{\left(\frac{1}{\text{AKS}}+\frac{1}{\text{ANS}}\right)}, &otherwise\end{cases}
	\end{split}\end{equation} 
	where AKS and ANS are the classification accuracy of known classes and the ACC of novel classes calculated by Eqs. \ref{eq:Acc} and \ref{eq:ACC}, respectively. Inspired by HNA, we harmonize AKS and ANS to yield HCA, and its higher score indicates more successful NCD or GCD. Its rationale and differences from ACC are elaborated in Appendix B. 
	
	\subsubsection{Implementation details}
	To simulate the open world scenarios, we randomly pick 4$/$40$/$80 classes from CIFAR10$/$CIFAR100$/$CUB as the initial known classes, and then randomly pick 2$/$20$/$40 classes from the remaining classes at each incremental step (3 steps in total). Each OWR phase requires sequential evaluation of the IL, OSR, and GCD performance of various methods with accuracy, HNA, and HCA as metrics, respectively. Specifically, the performance of IL can be evaluated using the current (novel known classes) and all previous (old classes) test sets. The performance of OSR can be evaluated using the current and all previous test sets (known classes) as well as the next training and test sets (unknown classes). Since OSR is an online process, the next training set that the model has not seen can also participate in the evaluation. Typically, we should evaluate the performance of GCD using all instances rejected by OSR. However, instances rejected by different methods are not identical. To be fair, we still only perform GCD on the rejected instances, but evaluate the performance of GCD with the same dataset as when evaluating OSR performance. Although this performance may be affected by the OSR results, convincing conclusions can still be drawn from comprehensive analysis and comparison. 
	
	For our method, ViT's DINO self-supervised pre-trained weights are provided by \cite{Caron}. For CIFAR10$/$CIFAR100$/$CUB, $|\mathcal{M}_r|$ is set to 20k$/$20k$/$2.4k. Considering that the default parameters in the original work \cite{Elhamifar} of DS3 have proven to be well inclusive, we keep the same configuration. To be lightweight, closed set classifiers are selected from MLP, SVM, and XGBoost with default parameters. The only hyperparameter $\alpha$ in OpenGCD is determined in $\{1e^{-10},\cdots,1e^{10}\}$ using the open set grid search protocol \cite{Mendes-O} with HNA as the criterion. ss-$k$-means++ is a non-parametric algorithm, and $|\mathcal{C}^{\max}_t|$ is set to a larger number, 500, for all datasets.
	
	For other methods, we employ the standard$/$open set grid search protocol with accuracy$/$HNA as the criterion to determine parameters about IL$/$OSR. $|\mathcal{M}_r|$ is the same setting as our method. Given that this work is the first attempt to assist OWR with GCD, we embed the proposed GCD approach into existing baselines to give them GCD capability. Likewise, $|\mathcal{C}^{\max}_t|$ is set to 500 for all datasets. As with our method, manual annotation is mandatory before the next phase begins.
	
	We implement our method using PyTorch 1.13.1 and run experiments on a RTX 3090 GPU. Our results are averaged over 5 runs for all datasets. 
	
	\subsection{Experimental results}
	\subsubsection{Comparison with the baselines}
	\begin{table}[!tp]
	\centering
	\caption{Comparison of OpenGCD with other OWR methods. Double dagger ($\ddagger$) denotes that the original method is unable to perform GCD, and we embed the proposed exemplar selection and GCD methods to give it GCD capability. Dagger ($\dagger$) denotes that the original method has its own exemplar selection mechanism, we only embed the proposed GCD method to give it GCD capability.}
	\label{tab:comparison} 
	\tabcolsep=0.5pt
	\resizebox{1\textwidth}{!}{ 
		\begin{tabular}{llllllllllllllllllllllllll}
			\toprule
			& & \multicolumn{8}{l}{\bf CIFAR10}& \multicolumn{8}{l}{\bf CIFAR100}& \multicolumn{8}{l}{\bf CUB}\\
			\cmidrule(lr){3-10}\cmidrule(lr){11-18}\cmidrule(lr){19-26}
			& & \multicolumn{5}{l}{\bf IL}& \bf OSR& \multicolumn{2}{l}{\bf GCD}& \multicolumn{5}{l}{\bf IL}& \bf OSR& \multicolumn{2}{l}{\bf GCD}& \multicolumn{5}{l}{\bf IL}& \bf OSR& \multicolumn{2}{l}{\bf GCD}\\
			\cmidrule(lr){3-7}\cmidrule(lr){8-8}\cmidrule(lr){9-10}\cmidrule(lr){11-15}\cmidrule(lr){16-16}\cmidrule(lr){17-18}\cmidrule(lr){19-23}\cmidrule(lr){24-24}\cmidrule(lr){25-26}
			\bf Phase& \bf Method& \multicolumn{5}{l}{\bf Acc}& \bf HNA& \bf Est. k& \bf HCA& \multicolumn{5}{l}{\bf Acc}& \bf HNA& \bf Est. k& \bf HCA& \multicolumn{5}{l}{\bf Acc}& \bf HNA& \bf Est. k& \bf HCA\\
			\midrule
			\multirow{5}*{\bf 1st} & & $\text{Ts}_\text{1}$& & & & $\text{Ts}_\text{1}$& $\text{Ts}_\text{1-2}\!+\!\text{Tr}_\text{2}$& $\text{k}_{\text{GT}}\!=\!6$& $\text{Ts}_\text{1-2}\!+\!\text{Tr}_\text{2}$& $\text{Ts}_\text{1}$& & & & $\text{Ts}_\text{1}$& $\text{Ts}_\text{1-2}\!+\!\text{Tr}_\text{2}$& $\text{k}_{\text{GT}}\!=\!60$& $\text{Ts}_\text{1-2}\!+\!\text{Tr}_\text{2}$& $\text{Ts}_\text{1}$& & & & $\text{Ts}_\text{1}$& $\text{Ts}_\text{1-2}\!+\!\text{Tr}_\text{2}$& $\text{k}_{\text{GT}}\!=\!120$& $\text{Ts}_\text{1-2}\!+\!\text{Tr}_\text{2}$\\
			& $^\dagger\text{NNO}$\cite{Bendale-15}& 87.9\%& & & & 87.9\%& 47.7\%& \bf 6& 25.4\%& 59.6\%& & & & 59.6\%& 48.5\%& 57& 16.4\%& 76.5\%& & & & 76.5\%& 60.4\%& \bf 120& 11.7\%\\
			& $^\ddagger\text{EVM}$\cite{Rudd}& 92.8\%& & & & 92.8\%& \bf 89.4\%& 7& 70.7\%& 71.6\%& & & & 71.6\%& 62.3\%& 56& 49.0\%& 44.3\%& & & & 44.3\%& 48.8\%& 144& 28.2\%\\
			& $^\dagger\text{L2AC}$\cite{Xu}& 93.8\%& & & & 93.8\%& 88.1\%& 7& 79.7\%& 59.1\%& & & & 59.1\%& 61.1\%& 67& 45.3\%& 65.9\%& & & & 65.9\%& 53.9\%& 138& 18.8\%\\
			& $^\dagger\text{DeepNNO}$\cite{Mancini}& 89.5\%& & & & 89.5\%& 47.5\%& 7& 23.1\%& 57.8\%& & & & 57.8\%& 41.5\%& \bf 60& 16.6\%& 82.5\%& & & & 82.5\%& 67.7\%& 118& 11.9\% \\
			& $^\dagger\text{B-DOC}$\cite{Fontanel}& 90.4\%& & & & 90.4\%& 85.7\%& 8& 40.4\%& 54.7\%& & & & 54.7\%& 60.9\%& 62& 24.9\%& 75.7\%& & & & 75.7\%& 57.8\%& 134& 17.1\%\\\cmidrule(lr){2-26}
			\rowcolor{mygray}\cellcolor{white}& $\text{OpenGCD}_\text{MLP}$& 98.6\%& & & & 98.6\%& 61.3\%& \bf 6& 73.7\%& 91.3\%& & & & 91.3\%& 72.6\%& 67& 56.5\%& 84.5\%& & & & 84.5\%& 64.2\%& 166& 22.8\%\\
			\rowcolor{mygray}\cellcolor{white}& $\text{OpenGCD}_\text{SVM}$& \bf 99.2\%& & & & \bf 99.2\%& 69.4\%& 7& 68.5\%& \bf 91.9\%& & & & \bf 91.9\%& \bf 85.2\%& 65& \bf 62.6\%& \bf 85.7\%& & & & \bf 85.7\%& \bf 70.4\%& 128& \bf 29.6\%\\
			\rowcolor{mygray}\cellcolor{white}& $\text{OpenGCD}_\text{XGB}$& 98.1\%& & & & 98.1\%& 83.0\%& \bf 6& \bf 85.1\%& 72.6\%& & & & 72.6\%& 67.3\%& 59& 54.2\%& 66.6\%& & & & 66.6\%& 58.7\%& \bf 120& 23.8\%\\
			\midrule
			\multirow{5}*{\bf 2nd} & & $\text{Ts}_\text{1}$& $\text{Ts}_\text{2}$& & & $\text{Ts}_\text{1-2}$& $\text{Ts}_\text{1-3}\!+\!\text{Tr}_\text{3}$& $\text{k}_{\text{GT}}\!=\!8$& $\text{Ts}_\text{1-3}\!+\!\text{Tr}_\text{3}$& $\text{Ts}_\text{1}$& $\text{Ts}_\text{2}$& & & $\text{Ts}_\text{1-2}$& $\text{Ts}_\text{1-3}\!+\!\text{Tr}_\text{3}$& $\text{k}_{\text{GT}}\!=\!80$& $\text{Ts}_\text{1-3}\!+\!\text{Tr}_\text{3}$& $\text{Ts}_\text{1}$& $\text{Ts}_\text{2}$& & & $\text{Ts}_\text{1-2}$& $\text{Ts}_\text{1-3}\!+\!\text{Tr}_\text{3}$& $\text{k}_{\text{GT}}\!=\!160$& $\text{Tst}_\text{1-3}\!+\!\text{Tr}_\text{3}$\\
			& $^\dagger\text{NNO}$\cite{Bendale-15}& 80.1\%& 84.9\%& & & 81.7\%& 50.5\%& 7& 17.9\%& 51.8\%& 51.7\%& & & 51.8\%& 43.2\%& 83& 16.1\%& 63.5\%& 60.5\%& & & 63.1\%& 52.5\%& 151& 9.1\%\\
			& $^\ddagger\text{EVM}$\cite{Rudd}& 86.6\%& 97.6\%& & & 89.8\%& \bf 84.9\%& \bf 8& \bf 91.7\%& 69.7\%& 68.3\%& & & 69.2\%& 68.9\%& \bf 79& 46.5\%& 36.3\%& 43.5\%& & & 37.3\%& 55.3\%& \bf 158& 5.0\%\\
			& $^\dagger\text{L2AC}$\cite{Xu}& 88.1\%& 87.4\%& & & 87.9\%& 81.3\%& \bf 8& 88.3\%& 51.9\%& 53.3\%& & & 52.1\%& 57.8\%& 78& 25.6\%& 61.6\%& 80.1\%& & & 64.5\%& 47.6\%& \bf 162& 11.2\%\\
			& $^\dagger\text{DeepNNO}$\cite{Mancini}& 83.8\%& 84.7\%& & & 84.1\%& 62.0\%& 9& 26.0\%& 57.4\%& 26.8\%& & & 47.1\%& 36.9\%& \bf 79& 16.0\%& 64.6\%& 60.2\%& & & 64.0\%& 56.2\%& 168& 11.4\%\\
			& $^\dagger\text{B-DOC}$\cite{Fontanel}& 86.0\%& 84.7\%& & & 85.5\%& 79.9\%& 9& 54.7\%& 50.0\%& 52.7\%& & & 51.0\%& 47.9\%& \bf 79& 20.9\%& 51.7\%& 53.5\%& & & 51.9\%& 48.9\%& 176& \bf 13.5\%\\\cmidrule(lr){2-26}
			\rowcolor{mygray}\cellcolor{white}& $\text{OpenGCD}_\text{MLP}$& 97.1\%& 96.1\%& & & 96.7\%& 84.6\%& \bf 8& 88.3\%& 88.0\%& 86.2\%& & & 87.4\%& \bf 80.2\%& 77& 45.2\%& \bf 79.3\%& \bf 80.3\%& & & \bf 79.5\%& 64.4\%& 175& 10.6\%\\
			\rowcolor{mygray}\cellcolor{white}& $\text{OpenGCD}_\text{SVM}$& \bf 98.6\%& \bf 98.9\%& & & \bf 98.7\%& 82.7\%& \bf 8& \bf 91.7\%& \bf 88.3\%& \bf 86.5\%& & & \bf 87.7\%& \bf 80.2\%& 77& 50.5\%& 79.0\%& 77.9\%& & & 78.8\%& \bf 66.1\%& 154& 7.3\%\\
			\rowcolor{mygray}\cellcolor{white}& $\text{OpenGCD}_\text{XGB}$& 95.5\%& 94.0\%& & & 95.0\%& 84.2\%& \bf 8& 89.4\%& 62.5\%& 59.8\%& & & 61.6\%& 65.5\%& 85& \bf 53.9\%& 57.6\%& 49.5\%& & & 56.4\%& 59.5\%& 145& 7.6\%\\
			\midrule
			\multirow{5}*{\bf 3rd} & & $\text{Ts}_\text{1}$& $\text{Ts}_\text{2}$& $\text{Ts}_\text{3}$& & $\text{Ts}_\text{1-3}$& $\text{Ts}_\text{1-4}\!+\!\text{Tr}_\text{4}$& $\text{k}_{\text{GT}}\!=\!10$& $\text{Ts}_\text{1-4}\!+\!\text{Tr}_\text{4}$& $\text{Ts}_\text{1}$& $\text{Ts}_\text{2}$& $\text{Ts}_\text{3}$& & $\text{Ts}_\text{1-3}$& $\text{Ts}_\text{1-4}\!+\!\text{Tr}_\text{4}$& $\text{k}_{\text{GT}}\!=\!100$& $\text{Ts}_\text{1-4}\!+\!\text{Tr}_\text{4}$& $\text{Ts}_\text{1}$& $\text{Ts}_\text{2}$& $\text{Ts}_\text{3}$& & $\text{Ts}_\text{1-3}$& $\text{Ts}_\text{1-4}\!+\!\text{Tr}_\text{4}$& $\text{k}_{\text{GT}}\!=\!200$& $\text{Ts}_\text{1-4}\!+\!\text{Tr}_\text{4}$\\
			& $^\dagger\text{NNO}$\cite{Bendale-15}& 82.0\%& 84.0\%& 89.5\%& & 84.4\%& 57.2\%& 9& 25.1\%& 45.3\%& 46.2\%& 47.3\%& & 46.0\%& 38.9\%& 92& 11.7\%& 40.3\%& 43.4\%& 44.4\%& & 41.2\%& 33.7\%& 194& 5.7\%\\
			& $^\ddagger\text{EVM}$\cite{Rudd}& 83.4\%& 85.3\%& 99.5\%& & 87.7\%& 79.8\%& \bf 10& 94.1\%& 48.5\%& 54.3\%& 53.9\%& & 51.3\%& 40.6\%& 98& 22.9\%& 15.4\%& 21.4\%& 11.9\%& & 15.7\%& 43.7\%& 183& 12.7\%\\
			& $^\dagger\text{L2AC}$\cite{Xu}& 86.0\%& 82.7\%& 90.2\%& & 86.2\%& 79.1\%& 9& 60.6\%& 48.1\%& 48.5\%& 48.6\%& & 48.2\%& 55.3\%& 99& 28.6\%& 56.7\%& \bf 79.2\%& 75.3\%& & 62.4\%& 52.9\%& \bf 203& 22.9\%\\
			& $^\dagger\text{DeepNNO}$\cite{Mancini}& 87.6\%& 80.3\%& 83.4\%& & 84.7\%& 52.6\%& \bf 10& 18.8\%& 55.7\%& 27.5\%& 19.9\%& & 39.6\%& 34.9\%& 90& 9.6\%& 29.2\%& 32.4\%& 23.7\%& & 28.9\%& 31.8\%& 210& 10.0\%\\
			& $^\dagger\text{B-DOC}$\cite{Fontanel}& 86.5\%& 82.8\%& 90.3\%& & 86.5\%& 48.8\%& \bf 10& 47.3\%& 47.1\%& 50.7\%& 49.8\%& & 48.7\%& 57.9\%& 92& 17.2\%& 27.4\%& 31.8\%& 27.3\%& & 27.9\%& 27.6\%& 224& 13.2\%\\\cmidrule(lr){2-26}
			\rowcolor{mygray}\cellcolor{white}& $\text{OpenGCD}_\text{MLP}$& 96.0\%& 95.7\%& 97.7\%& & 96.3\%& \bf 81.8\%& \bf 10& 89.8\%& 84.7\%& 82.7\%& 84.3\%& & 84.1\%& 74.5\%& 97& 30.2\%& \bf 74.5\%& 74.9\%& 76.4\%& & \bf 74.8\%& 60.7\%& 222& 21.9\%\\
			\rowcolor{mygray}\cellcolor{white}& $\text{OpenGCD}_\text{SVM}$& \bf 98.1\%& \bf 98.6\%& \bf 99.6\%& & \bf 98.5\%& 77.4\%& \bf 10& \bf 94.5\%& \bf 85.9\%& \bf 84.9\%& \bf 85.1\%& & \bf 85.4\%& \bf 75.1\%& 95& 38.9\%& 74.0\%& 75.4\%& \bf 76.9\%& & 74.6\%& \bf 65.4\%& 214& \bf 32.0\%\\
			\rowcolor{mygray}\cellcolor{white}& $\text{OpenGCD}_\text{XGB}$& 94.5\%& 92.5\%& 95.3\%& & 94.2\%& 79.1\%& \bf 10& 91.6\%& 57.1\%& 56.1\%& 54.6\%& & 56.2\%& 61.8\%& \bf 100& \bf 40.6\%& 48.0\%& 38.6\%& 41.0\%& & 45.9\%& 56.3\%& 178& 16.7\%\\
			\midrule
			\multirow{5}*{\bf 4th} & & $\text{Ts}_\text{1}$& $\text{Ts}_\text{2}$& $\text{Ts}_\text{3}$& $\text{Ts}_\text{4}$& $\text{Ts}_\text{1-4}$ &&&& $\text{Ts}_\text{1}$& $\text{Ts}_\text{2}$& $\text{Ts}_\text{3}$& $\text{Ts}_\text{4}$& $\text{Ts}_\text{1-4}$ &&&& $\text{Ts}_\text{1}$& $\text{Ts}_\text{2}$& $\text{Ts}_\text{3}$& $\text{Ts}_\text{4}$& $\text{Ts}_\text{1-4}$ &&&\\
			& $^\dagger\text{NNO}$\cite{Bendale-15}& 83.8\%& 63.0\%& 85.2\%& 90.7\%& 81.3\%& & & & 28.4\%& 37.3\%& 34.2\%& 28.4\%& 31.4\%& & && 35.9\%& 34.9\%& 36.4\%& 41.2\%& 36.4\%& & &\\
			& $^\ddagger\text{EVM}$\cite{Rudd}& 77.1\%& 85.4\%& 98.4\%& 97.8\%& 86.9\%& & & & 49.4\%& 54.3\%& 49.8\%& 60.8\%& 52.8\%& & && 12.7\%& 19.4\%& 10.7\%& 29.7\%& 15.2\%& & &\\
			& $^\dagger\text{L2AC}$\cite{Xu}& 83.5\%& 82.7\%& 90.1\%& 93.3\%& 86.6\%& & & & 44.3\%& 45.4\%& 45.7\%& 49.2\%& 45.0\%& & && 55.1\%& \bf 78.3\%& \bf 75.3\%& 87.4\%& 64.3\%& & &\\
			& $^\dagger\text{DeepNNO}$\cite{Mancini}& 78.0\%& 86.1\%& 87.4\%& 92.7\%& 84.5\%& & & & 53.3\%& 29.3\%& 22.7\%& 15.0\%& 34.6\%& & & & 14.4\%& 14.5\%& 13.9\%& 9.2\%& 13.7\%& & &\\
			& $^\dagger\text{B-DOC}$\cite{Fontanel}& 82.2\%& 76.1\%& 93.5\%& 93.0\%& 85.4\%& & & & 44.1\%& 46.5\%& 41.9\%& 43.0\%& 43.9\%& & & & 22.8\%& 23.6\%& 26.2\%& 24.1\%& 23.5\%& & &\\\cmidrule(lr){2-26}
			\rowcolor{mygray}\cellcolor{white}& $\text{OpenGCD}_\text{MLP}$& 94.9\%& 95.6\%& 97.7\%& 97.8\%& 96.2\%& & & & 82.1\%& 78.8\%& 81.1\%& 80.8\%& 81.0\%& & & & \bf 74.9\%& 72.6\%& 74.9\%& \bf 90.0\%& \bf 76.4\%& & &\\
			\rowcolor{mygray}\cellcolor{white}& $\text{OpenGCD}_\text{SVM}$& \bf 97.3\%& \bf 98.5\%& \bf 99.6\%& \bf 99.2\%& \bf 98.4\%& & & & \bf 83.6\%& \bf 82.6\%& \bf 83.4\%& \bf 82.9\%& \bf 83.2\%& & & & 74.4\%& 72.6\%& 74.4\%& 88.3\%& 75.8\%& & &\\
			\rowcolor{mygray}\cellcolor{white}& $\text{OpenGCD}_\text{XGB}$& 91.0\%& 91.3\%& 94.8\%& 96.0\%& 92.8\%& & & & 51.5\%& 49.9\%& 47.9\%& 51.4\%& 50.4\%& & & & 42.8\%& 40.3\%& 34.6\%& 55.1\%& 43.0\%& & &\\
			\bottomrule
	\end{tabular}}
	\parbox{0.99\textwidth}{\footnotesize\emph{We abbreviate the $i^{th}$ training$/$test set as $\text{Tr}_\text{i}/\text{Ts}_\text{i}$. In this work, this abbreviation will be followed subsequently.}}
	\end{table}
	We compare OpenGCD armed with MLP, SVM, and XGBoost against the state-of-the-art baselines for OWR, starting from CIFAR10, CIFAR100, and CUB in Tab. \ref{tab:comparison}. 
	
	For IL (columns 3-7, 11-15, and 19-23 in Tab. \ref{tab:comparison}), the best accuracies are almost all in the gray zones, which indicates that the proposed exemplar-based IL scheme substantially outperforms the other baselines. For column 7$/$15$/$23, the least decrease in average accuracy is found in $\text{OpenGCD}_{\text{SVM}}/\text{OpenGCD}_{\text{SVM}}/\text{L2AC}$ with 0.8\%$/$8.7\%$/$1.6\%, which indicates that the proposed IL scheme excels in both learning novel knowledge and retaining old knowledge. L2AC's close win on CUB indicates that trading time and space for performance is costly but effective in the case of small sample size and large class number. Longitudinally, the same method shows a decreasing trend in recognition ability for the same test set at different phases. This is the dual effect of increasing difficulty due to increasing number of classes and decreasing number of instances in each class due to constant buffer size. 
	
	For OSR (columns 8, 16, and 24 in Tab. \ref{tab:comparison}), the best HNAs are also concentrated in the gray zones, except for the first and second phases where EVM is slightly better on CIFAR10, which indicates that the proposed uncertainty-based OSR scheme significantly outperforms the other rivals. The proposed IL scheme's endeavor to avoid catastrophic forgetting and preserve the original spatial information are the magic bullet for OpenGCD to turn the tables. Moreover, the proposed OSR scheme is not only computationally lightweight, but also visualizes the probability distribution of instances over unknown and all known classes, which is not available in the other methods. We empirically found that HNA did not show a continuous downward trend over time, as accuracy did, but rather fluctuated downward. This is reasonable since while OSR performance is strongly dependent on model accuracy, it also plays a key role in whether the difference between unknown and known classes is significant.
	
	For GCD (columns 9-10, 17-18, and 25-26 in Tab. \ref{tab:comparison}), all results are generated using the proposed class number estimation and GCD schemes. The average estimation errors on the three datasets are 6.6\%, 4.5\%, and 8.0\%, respectively, which indicates the effectiveness of the fine-tuned class number estimation protocol. It is worth mentioning that Brent's algorithm converges at most in the $12^{th}$ epoch (occurring at the second phase of $\text{OpenGCD}_{\text{MLP}}$ on CUB), which improves the search efficiency by 30.7 times compared to the original protocol. Almost all the best HCAs are also located within the gray zones, benefiting from the excellent performance of the proposed IL and OSR schemes. If a method is slightly inferior on HNA but catches up on HCA, it means that the method offers a better fit with the proposed GCD, such as $\text{OpenGCD}_{\text{XGB}}$, which is at the third phase on CIFAR100. The comparison of EVM with NNO, DeepNNO and B-DOC reveals the importance of the exemplar selection strategy. The latter three focus excessively on representativeness rather than diversity of exemplars, resulting in inadequate retention of original information and hence poor performance. L2AC has a silver lining only by virtue of its multiple utilization of exemplars. Although the performance of the proposed GCD scheme is unsatisfactory in the case of a large number of classes, the results still demonstrate the feasibility of the attempt to assist OWR with GCD.
	
	Overall, almost all best performance is concentrated in the gray zones, which well demonstrates the technical advancement and excellent compatibility of OpenGCD.
	
	\subsubsection{Ablation study}
	\begin{table}[!tbp]
		\centering
		\caption{Ablation study of OpenGCD.}
		\label{tab:ablation} 
		\tabcolsep=1pt
		\resizebox{1\textwidth}{!}{ 
			\begin{tabular}{lllllllllllllllllllllll}
				\toprule
				& & \multicolumn{7}{l}{\bf CIFAR10}& \multicolumn{7}{l}{\bf CIFAR100}& \multicolumn{7}{l}{\bf CUB}\\
				\cmidrule(lr){3-9}\cmidrule(lr){10-16}\cmidrule(lr){17-23}
				& & \multicolumn{5}{l}{\bf IL}& \bf OSR& \bf GCD& \multicolumn{5}{l}{\bf IL}& \bf OSR& \bf GCD& \multicolumn{5}{l}{\bf IL}& \bf OSR& \bf GCD\\
				\cmidrule(lr){3-7}\cmidrule(lr){8-8}\cmidrule(lr){9-9}\cmidrule(lr){10-14}\cmidrule(lr){15-15}\cmidrule(lr){16-16}\cmidrule(lr){17-21}\cmidrule(lr){22-22}\cmidrule(lr){23-23}
				\bf Phase& \bf Method& \multicolumn{5}{l}{\bf Acc}& \bf HNA& \bf HCA& \multicolumn{5}{l}{\bf Acc}& \bf HNA& \bf HCA& \multicolumn{5}{l}{\bf Acc}& \bf HNA& \bf HCA\\
				\midrule
				\multirow{5}*{\bf 1st} & & $\text{Ts}_\text{1}$& & & & $\text{Ts}_\text{1}$& $\text{Ts}_\text{1-2}\!+\!\text{Tr}_\text{2}$& $\text{Ts}_\text{1-2}\!+\!\text{Tr}_\text{2}$& $\text{Ts}_\text{1}$& & & & $\text{Ts}_\text{1}$& $\text{Ts}_\text{1-2}\!+\!\text{Tr}_\text{2}$& $\text{Ts}_\text{1-2}\!+\!\text{Tr}_\text{2}$& $\text{Ts}_\text{1}$& & & & $\text{Ts}_\text{1}$& $\text{Ts}_\text{1-2}\!+\!\text{Tr}_\text{2}$& $\text{Ts}_\text{1-2}\!+\!\text{Tr}_\text{2}$\\
				& $\text{CSR}_\text{XGB}$& \bf 98.1\%& & & & \bf 98.1\%& 0\%& 0\%& \bf 72.6\%& & & & \bf 72.6\%& 0\%& 0\%& \bf 66.6\%& & & & \bf 66.6\%& 0\%& 0\%\\
				& $\text{IL-E}_\text{XGB}$& \bf 98.1\%& & & & \bf 98.1\%& 0\%& 0\%& \bf 72.6\%& & & & \bf 72.6\%& 0\%& 0\%& \bf 66.6\%& & & & \bf 66.6\%& 0\%& 0\%\\
				& $\text{OWR-UE}_\text{XGB}$& \bf 98.1\%& & & & \bf 98.1\%& \bf 83.0\%& 49.5\%& \bf 72.6\%& & & & \bf 72.6\%& \bf 67.3\%& 8.5\%& \bf 66.6\%& & & & \bf 66.6\%& \bf 58.7\%& 4.7\%\\
				& $\text{OpenGCD}_\text{XGB}$& \bf 98.1\%& & & & \bf 98.1\%& \bf 83.0\%& \bf 85.1\%& \bf 72.6\%& & & & \bf 72.6\%& \bf 67.3\%& \bf 54.2\%& \bf 66.6\%& & & & \bf 66.6\%& \bf 58.7\%& \bf 23.8\%\\
				\midrule
				\multirow{5}*{\bf 2nd} & & $\text{Ts}_\text{1}$& $\text{Ts}_\text{2}$& & & $\text{Ts}_\text{1-2}$& $\text{Ts}_\text{1-3}\!+\!\text{Tr}_\text{3}$& $\text{Ts}_\text{1-3}\!+\!\text{Tr}_\text{3}$& $\text{Ts}_\text{1}$& $\text{Ts}_\text{2}$& & & $\text{Ts}_\text{1-2}$& $\text{Ts}_\text{1-3}\!+\!\text{Tr}_\text{3}$& $\text{Ts}_\text{1-3}\!+\!\text{Tr}_\text{3}$& $\text{Ts}_\text{1}$& $\text{Ts}_\text{2}$& & & $\text{Ts}_\text{1-2}$& $\text{Ts}_\text{1-3}\!+\!\text{Tr}_\text{3}$& $\text{Ts}_\text{1-3}\!+\!\text{Tr}_\text{3}$\\
				& $\text{CSR}_\text{XGB}$& \bf 98.1\%& 0\%& & & 65.4\%& 0\%& 0\%& \bf 72.6\%& 0\%& & & 48.4\%& 0\%& 0\%& \bf 66.6\%& 0\%& & & 44.3\%& 0\%& 0\%\\
				& $\text{IL-E}_\text{XGB}$& 95.0\%& \bf 94.4\%& & & 94.8\%& 0\%& 0\%& 62.5\%& 60.1\%& & & 61.7\%& 0\%& 0\%& 56.3\%& \bf 50.4\%& & & 54.3\%& 0\%& 0\%\\
				& $\text{OWR-UE}_\text{XGB}$& 95.4\%& 94.1\%& & & \bf 95.0\%& 84.0\%& 41.8\%& 63.9\%& \bf 61.3\%& & & \bf 63.1\%& \bf 68.8\%& 8.3\%& 58.7\%& 49.0\%& & & \bf 57.3\%& \bf 59.5\%& 4.6\%\\
				& $\text{OpenGCD}_\text{XGB}$& 95.5\%& 94.0\%& & & \bf 95.0\%& \bf 84.2\%& \bf 89.4\%& 62.5\%& 59.8\%& & & 61.6\%& 65.5\%& \bf 53.9\%& 57.6\%& 49.5\%& & & 56.4\%& \bf 59.5\%& \bf 7.6\%\\
				\midrule
				\multirow{5}*{\bf 3rd} & & $\text{Ts}_\text{1}$& $\text{Ts}_\text{2}$& $\text{Ts}_\text{3}$& & $\text{Ts}_\text{1-3}$& $\text{Ts}_\text{1-4}\!+\!\text{Tr}_\text{4}$& $\text{Ts}_\text{1-4}\!+\!\text{Tr}_\text{4}$& $\text{Ts}_\text{1}$& $\text{Ts}_\text{2}$& $\text{Ts}_\text{3}$& & $\text{Ts}_\text{1-3}$& $\text{Ts}_\text{1-4}\!+\!\text{Tr}_\text{4}$& $\text{Ts}_\text{1-4}\!+\!\text{Tr}_\text{4}$& $\text{Ts}_\text{1}$& $\text{Ts}_\text{2}$& $\text{Ts}_\text{3}$& & $\text{Ts}_\text{1-3}$& $\text{Ts}_\text{1-4}\!+\!\text{Tr}_\text{4}$& $\text{Ts}_\text{1-4}\!+\!\text{Tr}_\text{4}$\\
				& $\text{CSR}_\text{XGB}$& \bf 98.1\%& 0\%& 0\%& & 49.0\%& 0\%& 0\%& \bf 72.6\%& 0\%& 0\%& & 36.3\%& 0\%& 0\%& \bf 66.6\%& 0\%& 0\%& & 33.2\%& 0\%& 0\%\\
				& $\text{IL-E}_\text{XGB}$& 93.9\%& 91.8\%& 95.2\%& & 93.7\%& 0\%& 0\%& 56.2\%& 54.7\%& 52.9\%& & 55.0\%& 0\%& 0\%& 48.1\%& 42.2\%& 37.7\%& & 44.0\%& 0\%& 0\%\\
				& $\text{OWR-UE}_\text{XGB}$& 93.4\%& 91.4\%& \bf 95.4\%& & 93.4\%& \bf 79.7\%& 38.5\%& 57.3\%& 53.3\%& \bf 54.7\%& & 55.7\%& 60.0\%& 8.1\%& 49.1\%& \bf 45.0\%& \bf 42.5\%& & \bf 47.7\%& \bf 57.5\%& 4.4\%\\
				& $\text{OpenGCD}_\text{XGB}$& 94.5\%& \bf 92.5\%& 95.3\%& & \bf 94.2\%& 79.1\%& \bf 91.6\%& 57.1\%& \bf 56.1\%& 54.6\%& & \bf 56.2\%& \bf 61.8\%& \bf 40.6\%& 48.0\%& 38.6\%& 41.0\%& & 45.9\%& 56.3\%& \bf 16.7\%\\
				\midrule
				\multirow{5}*{\bf 4th} & & $\text{Ts}_\text{1}$& $\text{Ts}_\text{2}$& $\text{Ts}_\text{3}$& $\text{Ts}_\text{4}$& $\text{Ts}_\text{1-4}$ & & & $\text{Ts}_\text{1}$& $\text{Ts}_\text{2}$& $\text{Ts}_\text{3}$& $\text{Ts}_\text{4}$& $\text{Ts}_\text{1-4}$ & & & $\text{Ts}_\text{1}$& $\text{Ts}_\text{2}$& $\text{Ts}_\text{3}$& $\text{Ts}_\text{4}$& $\text{Ts}_\text{1-4}$ & &\\
				& $\text{CSR}_\text{XGB}$& \bf 98.1\%& 0\%& 0\%& 0\%& 39.2\%& & & \bf 72.6\%& 0\%& 0\%& 0\%& 29.0\%& & & \bf 66.6\%& 0\%& 0\%& 0\%& 26.6\%& &\\
				& $\text{IL-E}_\text{XGB}$& 91.1\%& 90.6\%& 94.2\%& 95.0\%& 92.4\%& & & 51.7\%& \bf 50.2\%& 46.9\%& 51.1\%& 50.3\%& & & 39.8\%& 38.8\%& 32.0\%& 53.9\%& 40.9\%& &\\
				& $\text{OWR-UE}_\text{XGB}$& 91.5\%& \bf 91.3\%& 94.5\%& 95.4\%& \bf 92.9\%& & & 52.1\%& 48.6\%& \bf 49.9\%& \bf 51.9\%& \bf 50.9\%& & & 42.8\%& \bf 41.3\%& \bf 35.8\%& \bf 55.4\%& \bf 43.3\%& &\\
				& $\text{OpenGCD}_\text{XGB}$& 91.0\%& \bf 91.3\%& \bf 94.8\%& \bf 96.0\%& 92.8\%& & & 51.5\%& 49.9\%& 47.9\%& 51.4\%& 50.4\%& & & 42.8\%& 40.3\%& 34.6\%& 55.1\%& 43.0\%& &\\
				\bottomrule
		\end{tabular}}
		\parbox{0.99\textwidth}{\footnotesize\emph{CSR: A rudimentary version without IL (Secs. \ref{sec:es} \& \ref{sec:il}), OSR (Sec. \ref{sec:OSR}), and GCD (Sec. \ref{sec:gcd}) capabilities. IL-E: A half-baked version without OSR (Sec. \ref{sec:OSR}) and GCD (Sec. \ref{sec:gcd}) capabilities. OWR-UE: A base version without GCD (Sec. \ref{sec:gcd}) capability. OpenGCD: The full version of the proposed method described in Sec. \ref{sec:OpenGCD}.}}
	\end{table}
	
	We inspect the contributions of the various components of OpenGCD. Given that OpenGCDs armed with various classifiers all exhibit similar variations, we only present the ablation results for the more efficient $\text{OpenGCD}_\text{XGB}$ in Tab. \ref{tab:ablation}. As we can see, all components contribute significantly, and removing any of them can result in significant performance degradation or even loss of functionality. The reason why OWR-UE still scores a little on HCA although it lacks GCD capability lies in the fact that it classifies all novel classes as unknown, which is equivalent to clustering into one class. The slightly inferior performance of OpenGCD for IL, especially on the latter two datasets, is due to the fact that OWR-UE is labor-intensive to label the data one by one, while OpenGCD only corrects for clusters of instances recognized as novel classes. Compared to the OWR-UE, the IL-E completely loses its OSR capability. This is a nightmare for an online recognition system towards the open world, as it cannot detect anomalies or isolate foreign intrusions promptly. The CSR that lost its IL capability maintains a consistent knowledge of $\text{Ts}_\text{1}$, which allowed it to stay well ahead on $\text{Ts}_\text{1}$ at different phases. At the first phase, there is no difference between CSR (training on the full training set) and the other three methods for IL performance, which further indicates the appropriateness of targeting diversity for exemplar selection in response to catastrophic forgetting.
	
	We analyse the effects of $|\mathcal{M}_r|$ and $\alpha$ on performance in Appendix C. 
	
	\section{Conclusion}
	In this paper, we proposed OpenGCD to address the three main tasks in OWR by combining a few new ideas. Firstly, we rejected the unknown based on the uncertainty of the classifier's prediction, which is lightweight and intuitive. Secondly, we clustered unlabeled unknown instances using ss-$k$-means++, which is the first attempt to assist manual grouping in OWR with GCD techniques driving OWR a small step closer to automation. Besides, we fine-tuned an existing class number evaluation protocol, which achieves efficiency gains using optimization instead of traversal. Further, we proposed a new metric called HCA to evaluate the performance of GCD, which achieves more reasonable results in a harmonic fashion. Finally, we selected informative exemplars with the goal of diversity to ensure smooth implementation of IL and GCD. 
	
	Remarkably, all procedures in OpenGCD are independent of the classifier type, which gives it excellent compatibility, \ie, it opens the gate towards the open world for any well-designed closed set classifier. Moreover, OpenGCD is also extremely scalable, and its OWR performance can be further improved by introducing classifier calibration technology, more advanced semi-supervised clustering and classification models, memory management strategies, etc. We consider the implementation of OWR in limited data scenarios, such as few-shot OWR, and its further automation as potential future research directions.
	
	{\small
		\bibliographystyle{unsrtnat}
		\bibliography{egbib}

\begin{thebibliography}{33}
\providecommand{\natexlab}[1]{#1}
\providecommand{\url}[1]{\texttt{#1}}
\expandafter\ifx\csname urlstyle\endcsname\relax
  \providecommand{\doi}[1]{doi: #1}\else
  \providecommand{\doi}{doi: \begingroup \urlstyle{rm}\Url}\fi

\bibitem[Bendale and Boult(2015)]{Bendale-15}
Abhijit Bendale and Terrance Boult.
\newblock Towards open world recognition.
\newblock In \emph{CVPR}, pages 1893--1902, 2015.

\bibitem[Rudd et~al.(2018)Rudd, Jain, Scheirer, and Boult]{Rudd}
Ethan~M. Rudd, Lalit~P. Jain, Walter~J. Scheirer, and Terrance~E. Boult.
\newblock The extreme value machine.
\newblock \emph{IEEE TPAMI}, 40\penalty0 (3):\penalty0 762--768, 2018.

\bibitem[Xu et~al.(2019)Xu, Liu, Shu, and Yu]{Xu}
Hu~Xu, Bing Liu, Lei Shu, and P.~Yu.
\newblock Open-world learning and application to product classification.
\newblock In \emph{The World Wide Web Conference}, pages 3413--3419, 2019.

\bibitem[Mancini et~al.(2019)Mancini, Karaoguz, Ricci, Jensfelt, and
  Caputo]{Mancini}
Massimiliano Mancini, Hakan Karaoguz, Elisa Ricci, Patric Jensfelt, and Barbara
  Caputo.
\newblock Knowledge is never enough: Towards web aided deep open world
  recognition.
\newblock In \emph{ICRA}, pages 9537--9543, 2019.

\bibitem[Fontanel et~al.(2020)Fontanel, Cermelli, Mancini, Buló, Ricci, and
  Caputo]{Fontanel}
Dario Fontanel, Fabio Cermelli, Massimiliano Mancini, Samuel~Rota Buló, Elisa
  Ricci, and Barbara Caputo.
\newblock Boosting deep open world recognition by clustering.
\newblock \emph{IEEE Robot. Autom. Let.}, 5\penalty0 (4):\penalty0 5985--5992,
  2020.

\bibitem[Geng et~al.(2021)Geng, Huang, and Chen]{Geng}
Chuanxing Geng, Shengjun Huang, and Songcan Chen.
\newblock Recent advances in open set recognition: A survey.
\newblock \emph{IEEE TPAMI}, 43\penalty0 (10):\penalty0 3614--3631, 2021.

\bibitem[Yoder and Priebe(2016)]{Jordan}
Jordan Yoder and Carey~E. Priebe.
\newblock Semi-supervised k-means++.
\newblock \emph{J. Stat. Comput. Sim.}, 87:\penalty0 2597--2608, 2016.

\bibitem[Han et~al.(2022)Han, Rebuffi, Ehrhardt, Vedaldi, and Zisserman]{Han-A}
Kai Han, Sylvestre-Alvise Rebuffi, Sébastien Ehrhardt, Andrea Vedaldi, and
  Andrew Zisserman.
\newblock Autonovel: Automatically discovering and learning novel visual
  categories.
\newblock \emph{IEEE TPAMI}, 44\penalty0 (10):\penalty0 6767--6781, 2022.

\bibitem[Vaze et~al.(2022)Vaze, Han, Vedaldi, and Zisserman]{Vaze}
Sagar Vaze, Kai Han, Andrea Vedaldi, and Andrew Zisserman.
\newblock Generalized category discovery.
\newblock In \emph{CVPR}, pages 7482--7491, 2022.

\bibitem[Han et~al.(2019)Han, Vedaldi, and Zisserman]{Han-L}
Kai Han, Andrea Vedaldi, and Andrew Zisserman.
\newblock Learning to discover novel visual categories via deep transfer
  clustering.
\newblock In \emph{ICCV}, pages 8400--8408, 2019.

\bibitem[Zhang et~al.(2022)Zhang, Jiang, Feng, Wu, Zhao, Wan, Tang, Jin, and
  Gao]{Zhang}
Xinwei Zhang, Jianwen Jiang, Yutong Feng, Zhi-Fan Wu, Xibin Zhao, Hai Wan,
  Mingqian Tang, Rong Jin, and Yue Gao.
\newblock Grow and merge: A unified framework for continuous categories
  discovery.
\newblock In \emph{NIPS}, volume~35, pages 27455--27468, 2022.

\bibitem[Zhao and Han(2021)]{Zhao}
Bingchen Zhao and Kai Han.
\newblock Novel visual category discovery with dual ranking statistics and
  mutual knowledge distillation.
\newblock In \emph{NIPS}, volume~34, pages 22982--22994, 2021.

\bibitem[Cao et~al.(2022)Cao, Brbi\'c, and Leskovec]{Cao}
Kaidi Cao, Maria Brbi\'c, and Jure Leskovec.
\newblock Open-world semi-supervised learning.
\newblock In \emph{ICLR}, 2022.

\bibitem[De~Lange et~al.(2022)De~Lange, Aljundi, Masana, Parisot, Jia,
  Leonardis, Slabaugh, and Tuytelaars]{DeLange-A}
Matthias De~Lange, Rahaf Aljundi, Marc Masana, Sarah Parisot, Xu~Jia, Aleš
  Leonardis, Gregory Slabaugh, and Tinne Tuytelaars.
\newblock A continual learning survey: Defying forgetting in classification
  tasks.
\newblock \emph{IEEE TPAMI}, 44\penalty0 (7):\penalty0 3366--3385, 2022.

\bibitem[Elhamifar et~al.(2016)Elhamifar, Sapiro, and Sastry]{Elhamifar}
Ehsan Elhamifar, Guillermo Sapiro, and S.~Shankar Sastry.
\newblock Dissimilarity-based sparse subset selection.
\newblock \emph{IEEE TPAMI}, 38\penalty0 (11):\penalty0 2182--2197, 2016.

\bibitem[Liu et~al.(2021)Liu, Schiele, and Sun]{Liu}
Yaoyao Liu, Bernt Schiele, and Qianru Sun.
\newblock Rmm: Reinforced memory management for class-incremental learning.
\newblock In \emph{NIPS}, volume~34, pages 3478--3490, 2021.

\bibitem[Wang et~al.(2022)Wang, Xu, Yang, He, Cao, and Huang]{Wang}
Zitai Wang, Qianqian Xu, Zhiyong Yang, Yuan He, Xiaochun Cao, and Qingming
  Huang.
\newblock Openauc: Towards auc-oriented open-set recognition.
\newblock In \emph{NIPS}, volume~35, pages 25033--25045, 2022.

\bibitem[Yang et~al.(2022)Yang, Wang, Zou, Zhou, Ding, Peng, Wang, Chen, Li,
  Sun, Du, Zhou, Zhang, Hendrycks, Li, and Liu]{Yang}
Jingkang Yang, Pengyun Wang, Dejian Zou, Zitang Zhou, Kunyuan Ding, Wenxuan
  Peng, Haoqi Wang, Guangyao Chen, Bo~Li, Yiyou Sun, Xuefeng Du, Kaiyang Zhou,
  Wayne Zhang, Dan Hendrycks, Yixuan Li, and Ziwei Liu.
\newblock Openood: Benchmarking generalized out-of-distribution detection.
\newblock In \emph{NIPS}, volume~35, pages 32598--32611, 2022.

\bibitem[Scheirer et~al.(2013)Scheirer, de~Rezende~Rocha, Sapkota, and
  Boult]{Scheirer}
Walter~J. Scheirer, Anderson de~Rezende~Rocha, Archana Sapkota, and Terrance~E.
  Boult.
\newblock Toward open set recognition.
\newblock \emph{IEEE TPAMI}, 35\penalty0 (7):\penalty0 1757--1772, 2013.

\bibitem[Ribeiro Mendes~Júnior et~al.(2022)Ribeiro Mendes~Júnior, Boult,
  Wainer, and Rocha]{Mendes-O}
Pedro Ribeiro Mendes~Júnior, Terrance~E. Boult, Jacques Wainer, and Anderson
  Rocha.
\newblock Open-set support vector machines.
\newblock \emph{IEEE Trans. Syst. Man Cy.-S.}, 52\penalty0 (6):\penalty0
  3785--3798, 2022.

\bibitem[Jain et~al.(2014)Jain, Scheirer, and Boult]{Jain}
Lalit~P. Jain, Walter~J. Scheirer, and Terrance~E. Boult.
\newblock Multi-class open set recognition using probability of inclusion.
\newblock In \emph{ECCV}, pages 393--409, 2014.

\bibitem[Mendes~J\'{u}nior et~al.(2017)Mendes~J\'{u}nior, Souza, Werneck,
  Stein, Pazinato, Almeida, Penatti, Torres, and Rocha]{Mendes-N}
Pedro~R. Mendes~J\'{u}nior, Roberto~M. Souza, Rafael~De Werneck, Bernardo~V.
  Stein, Daniel~V. Pazinato, Waldir~R. Almeida, Ot\'{a}vio~A. Penatti,
  Ricardo~Da Torres, and Anderson Rocha.
\newblock Nearest neighbors distance ratio open-set classifier.
\newblock \emph{Mach. Learn.}, 106:\penalty0 359--386, 2017.

\bibitem[Bendale and Boult(2016)]{Bendale-16}
Abhijit Bendale and Terrance~E. Boult.
\newblock Towards open set deep networks.
\newblock In \emph{CVPR}, pages 1563--1572, 2016.

\bibitem[Gao et~al.(2023)Gao, Peng, Yang, Su, Li, and Zhong]{Gao}
Fulin Gao, Xin Peng, Dan Yang, Cheng Su, Linlin Li, and Weimin Zhong.
\newblock A novel distributed fault diagnosis scheme toward open-set scenarios
  based on extreme value theory.
\newblock \emph{IEEE Trans. Ind. Inform.}, pages 1--13, 2023.

\bibitem[Sun et~al.(2022)Sun, Lyu, Shang, Feng, and Wan]{Sun}
Qing Sun, Fan Lyu, Fanhua Shang, Wei Feng, and Liang Wan.
\newblock Exploring example influence in continual learning.
\newblock In \emph{NIPS}, volume~35, pages 27075--27086, 2022.

\bibitem[Li and Hoiem(2018)]{Li}
Zhizhong Li and Derek Hoiem.
\newblock Learning without forgetting.
\newblock \emph{IEEE TPAMI}, 40\penalty0 (12):\penalty0 2935--2947, 2018.

\bibitem[Mallya and Lazebnik(2018)]{Mallya}
Arun Mallya and Svetlana Lazebnik.
\newblock Packnet: Adding multiple tasks to a single network by iterative
  pruning.
\newblock In \emph{CVPR}, pages 7765--7773, 2018.

\bibitem[De~Lange and Tuytelaars(2021)]{DeLange-C}
Matthias De~Lange and Tinne Tuytelaars.
\newblock Continual prototype evolution: Learning online from non-stationary
  data streams.
\newblock In \emph{ICCV}, pages 8230--8239, 2021.

\bibitem[Caron et~al.(2021)Caron, Touvron, Misra, Jegou, Mairal, Bojanowski,
  and Joulin]{Caron}
Mathilde Caron, Hugo Touvron, Ishan Misra, Hervé Jegou, Julien Mairal, Piotr
  Bojanowski, and Armand Joulin.
\newblock Emerging properties in self-supervised vision transformers.
\newblock In \emph{ICCV}, pages 9630--9640, 2021.

\bibitem[Kuhn(1955)]{Harold}
Harold~W. Kuhn.
\newblock The hungarian method for the assignment problem.
\newblock \emph{Nav. Res. Log.}, 52, 1955.

\bibitem[Krizhevsky(2009)]{Alex}
Alex Krizhevsky.
\newblock Learning multiple layers of features from tiny images.
\newblock Technical report, 2009.

\bibitem[Wah et~al.(2011)Wah, Branson, Welinder, Perona, and Belongie]{Wah}
C.~Wah, S.~Branson, P.~Welinder, P.~Perona, and S.~Belongie.
\newblock Caltech-ucsd birds-200-2011.
\newblock Technical Report CNS-TR-2011-001, California Institute of Technology,
  2011.

\bibitem[Deng et~al.(2009)Deng, Dong, Socher, Li, Li, and Fei-Fei]{Deng}
Jia Deng, Wei Dong, Richard Socher, Li-Jia Li, Kai Li, and Li~Fei-Fei.
\newblock Imagenet: A large-scale hierarchical image database.
\newblock In \emph{CVPR}, pages 248--255, 2009.

\end{thebibliography}
	}
	\clearpage
	\appendix
	\section{Schematic of OpenGCD}\label{sec:A}
\begin{figure*}[!htbp]
		\centering
		\includegraphics[width=1.0\linewidth]{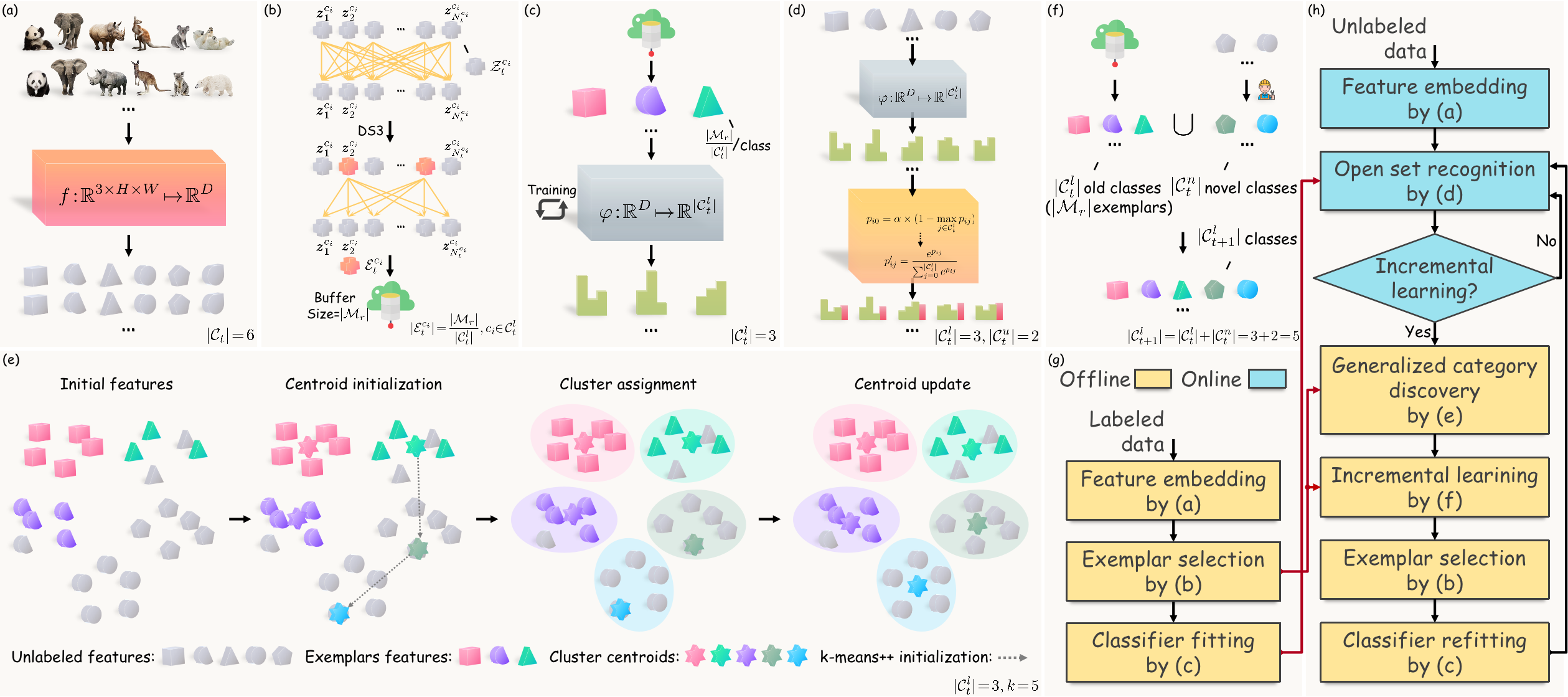}
		\caption{Schematic of OpenGCD. 
			(a) Feature embedding. No matter online or offline instance, its feature embedding should be obtained by ViT first (details c.f. Sec. 3.1 of the main paper). 
			(b) Exemplar selection. For the known class $c_i$, $|\mathcal{M}_r|/|\mathcal{C}_t^l|$ feature exemplars are selected by the DS3 algorithm to represent the original feature subset (details c.f. Sec. 3.2 of the main paper). 
			(c) Classifier (re)fitting. $|\mathcal{M}_r|$ exemplars from $\mathcal{C}_t^l$ are taken to (re)fit the classifier (details c.f. Sec. 3.3 of the main paper). 
			(d) Uncertainty-based OSR. For an online instance, the uncertainty of the closed set output of the classifier serves as an estimate for the unknown (details c.f. Sec. 3.4 of the main paper). 
			(e) ss-$k$-means++ algorithm for $k\!=\!5$. The centroids of $|\mathcal{C}_t^l|$ known classes are derived from partially labeled exemplars (Initial features). The centroids of the remaining $k\!-\!|\mathcal{C}_t^l|$ novel classes are initialized by $k$-means++ (Centroids initialization). For unlabeled features, clustering labels are assigned by identifying the nearest centroid (Cluster assignment). Centroids are updated by averaging the features in each cluster (Centroid update). Cluster assignment and Centroid update are then repeated until convergence, during which labeled exemplars are forced to follow their ground-truth labels (details c.f. Sec. 3.5.1 of the main paper). 
			(f) Exemplar-based IL. Merging the exemplar set $\mathcal{E}_t$ and the manually labeled novel class feature set $\mathcal{Z}_t^n$ yields the labeled feature set $\mathcal{Z}_{t+1}^l$ for IL (details c.f. Sec. 3.6 of the main paper). 
			(g) Offline modeling procedures. Pre-stored exemplars and fitted classifier are prepared for subsequent procedures.
			(h) Assisting OWR procedures with GCD. Online instances can be input separately or in batches. Once a phase of OSR is completed, GCD and IL can be launched.}
		\label{fig:frame}
	\end{figure*}

	The schematic of the formulated OpenGCD (from Sec. 3 of the main paper) is shown in Fig. \ref{fig:frame}, where (g) and (h) are flowcharts of the proposed solution and (a)-(f) are descriptions of each component. 
	
	\section{Harmonic clustering accuracy}\label{sec:B}
\begin{figure}[!tbp]
		\centering
		\includegraphics[width=3.5in]{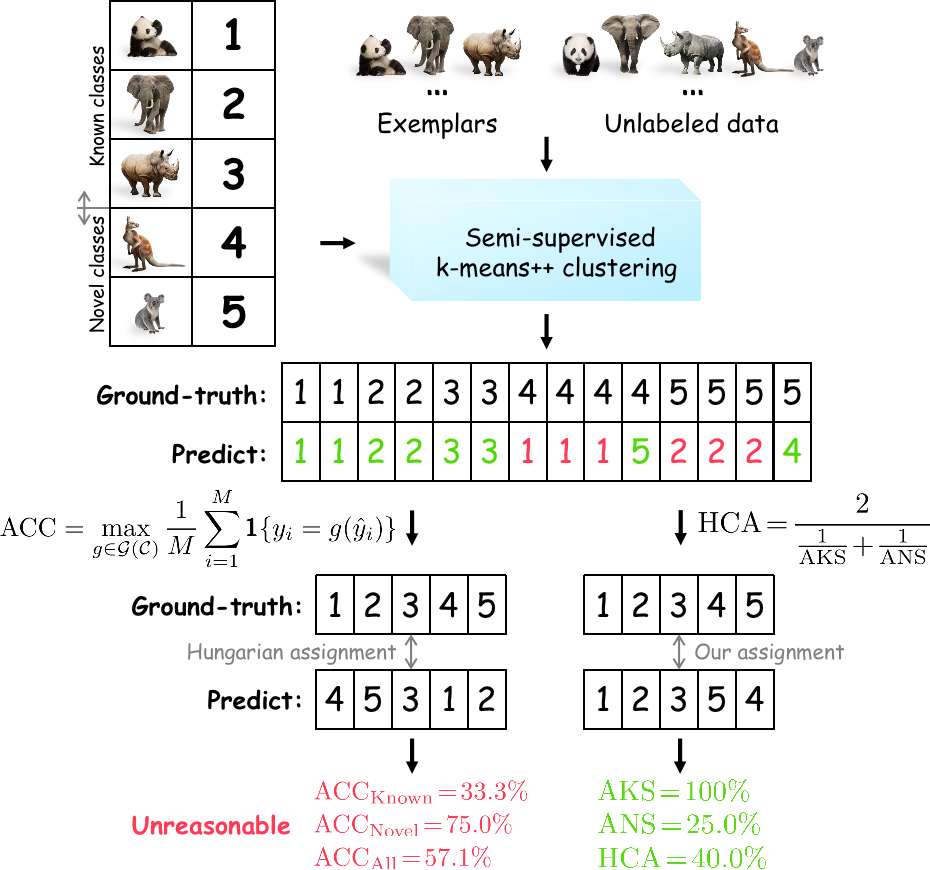}
		\caption{Schematic of ACC and HCA. In the case of obtaining the predicted results in the middle of the figure, ACC will match predicted clusters 4, 5 with ground-truth classes 1, 2, resulting in unreasonable evaluation of known and novel classes. In contrast, the proposed HCA evaluates known classes by classification accuracy (without matching) and novel classes by ACC (remaining clusters and ground-truth labels are matched by Hungarian algorithm), and finally harmonizes the two to yield more reasonable assignments and evaluations.}
		\label{fig:HCA}
	\end{figure}

	NCD or GCD is to classify known classes and cluster novel ones, so it is improper to do cluster assignment for known classes. However, as shown in Fig. \ref{fig:HCA}, ACC indiscriminately matches ground-truth and predicted labels in the greediest fashion, making it suitable only for clustering problems instead of NCD or GCD. With this motivation, we devised the HCA by imitating the HNA. As can be seen from Fig. \ref{fig:HCA}, HCA is more reasonable than ACC in evaluating NCD or GCD.
	
	\section{Parametric analysis}\label{sec:C}
	To avoid redundancy, we only analyze the parameters that need to be set manually in $\text{OpenGCD}_\text{XGB}$ with control variates.
	
\begin{figure}[!tbp]
		\centering
		\subfigure[]{\includegraphics[scale=0.28]{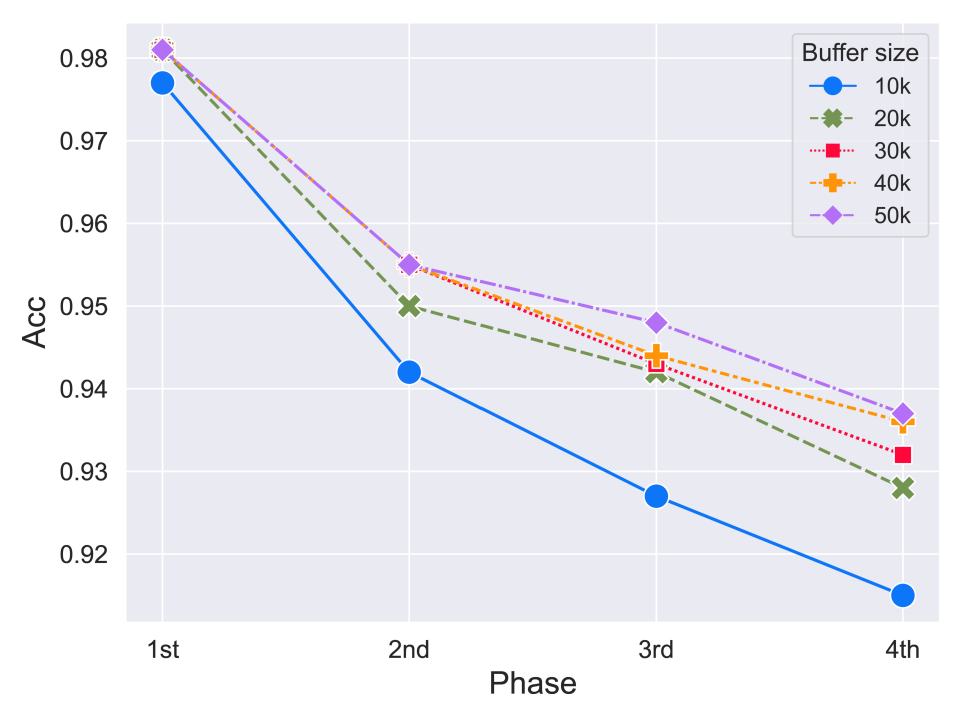}}
		\subfigure[]{\includegraphics[scale=0.28]{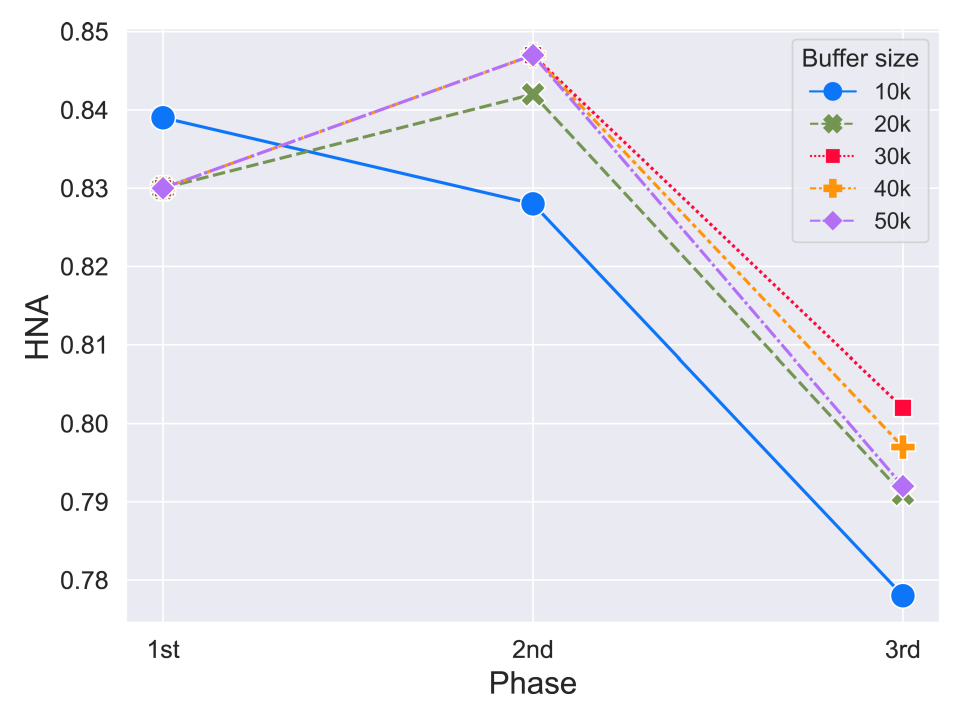}}
		\subfigure[]{\includegraphics[scale=0.28]{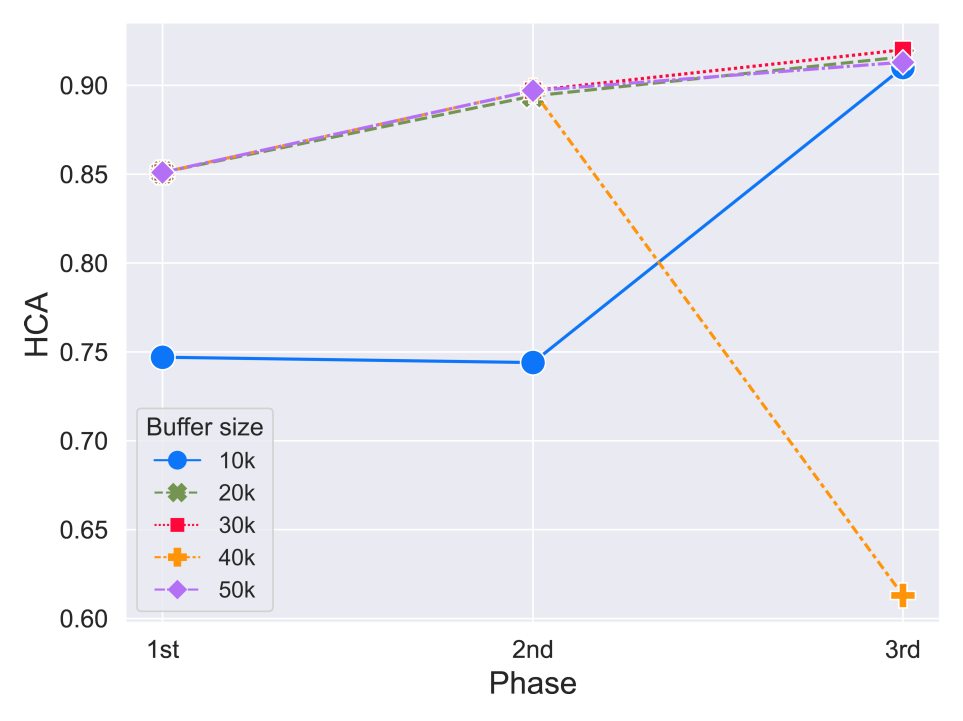}}
		\subfigure[]{\includegraphics[scale=0.28]{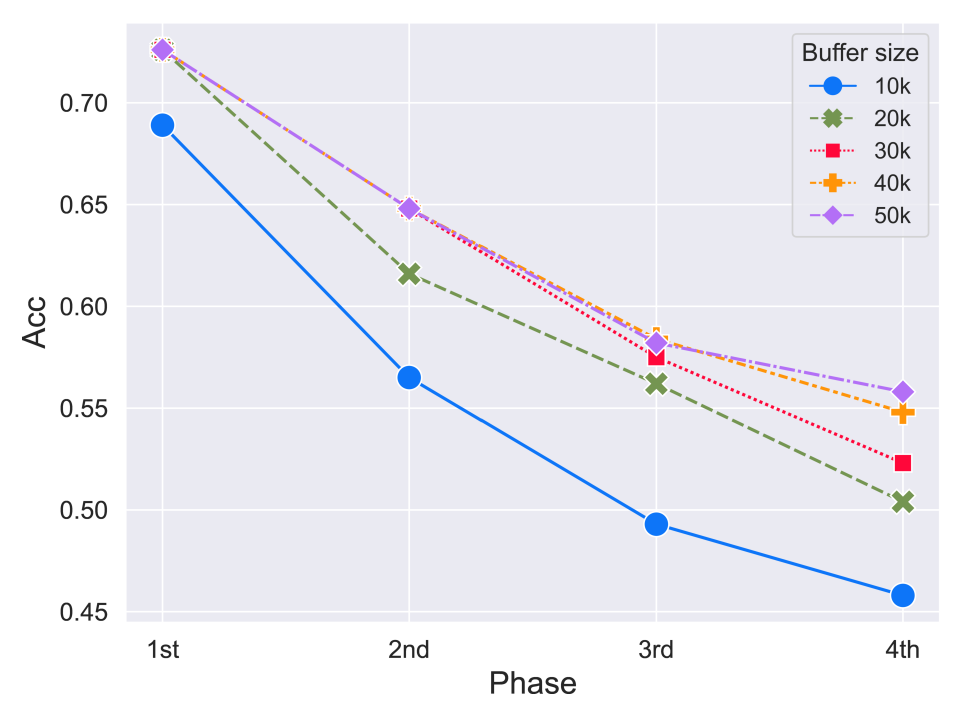}}
		\subfigure[]{\includegraphics[scale=0.28]{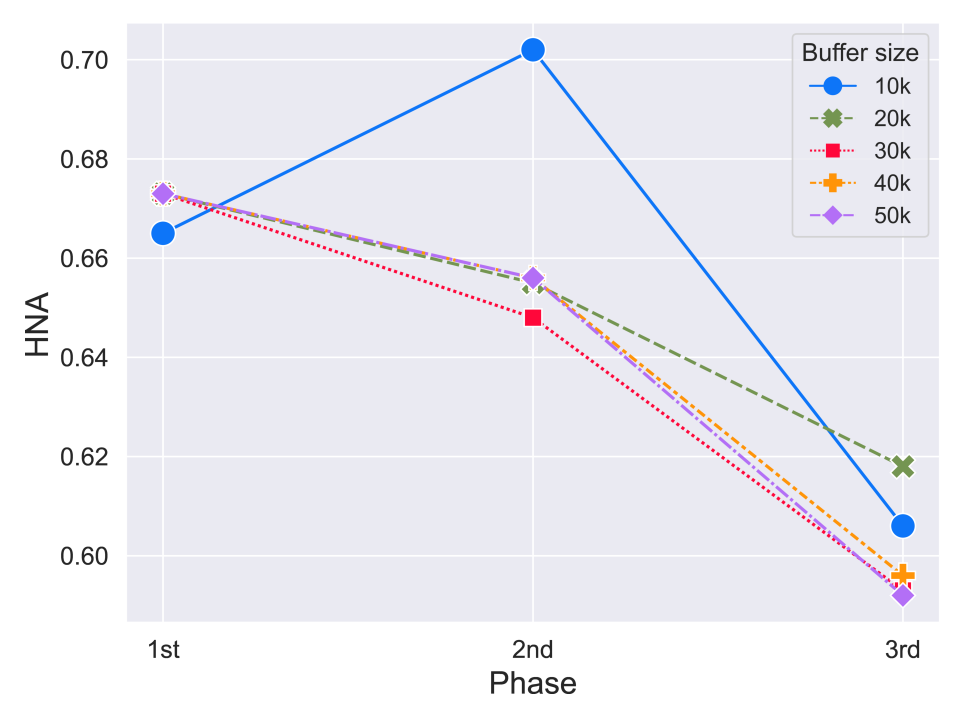}}
		\subfigure[]{\includegraphics[scale=0.28]{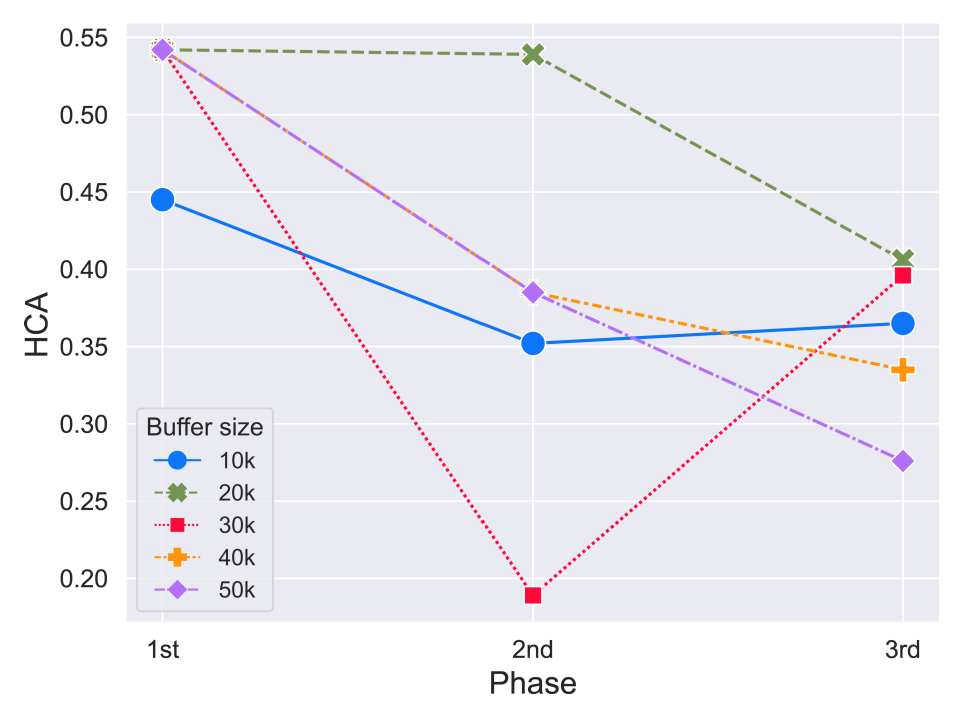}}
		\subfigure[]{\includegraphics[scale=0.28]{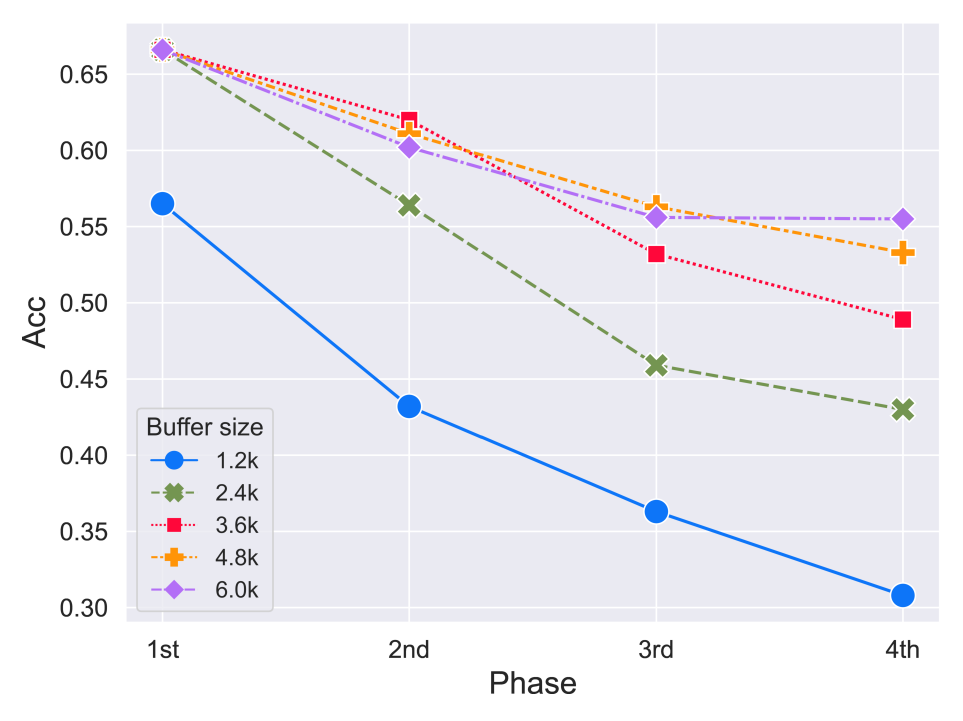}} 
		\subfigure[]{\includegraphics[scale=0.28]{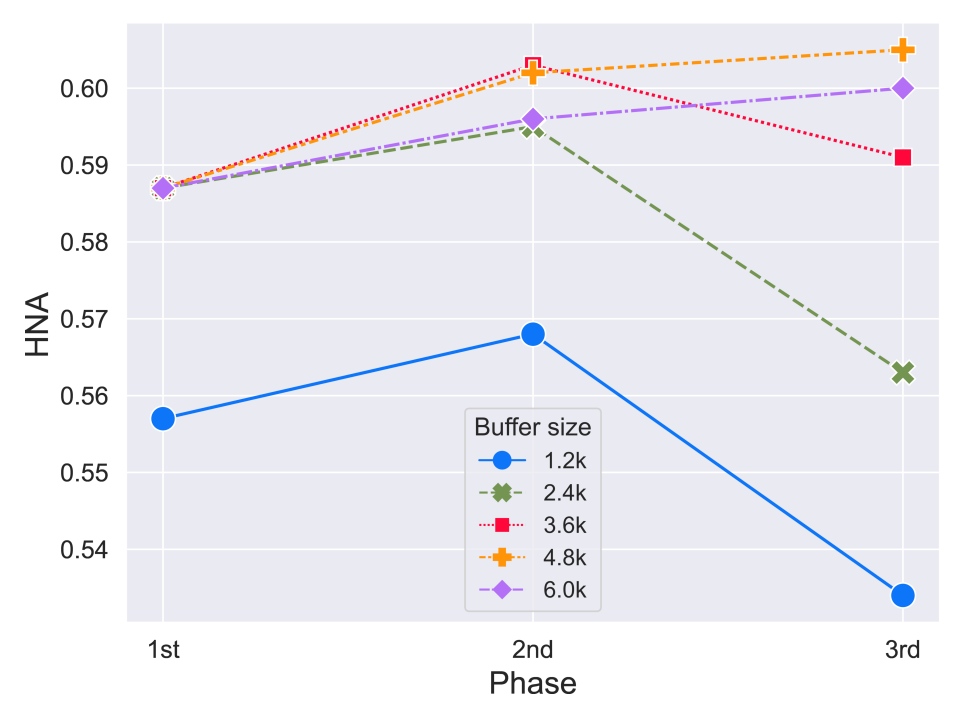}}
		\subfigure[]{\includegraphics[scale=0.28]{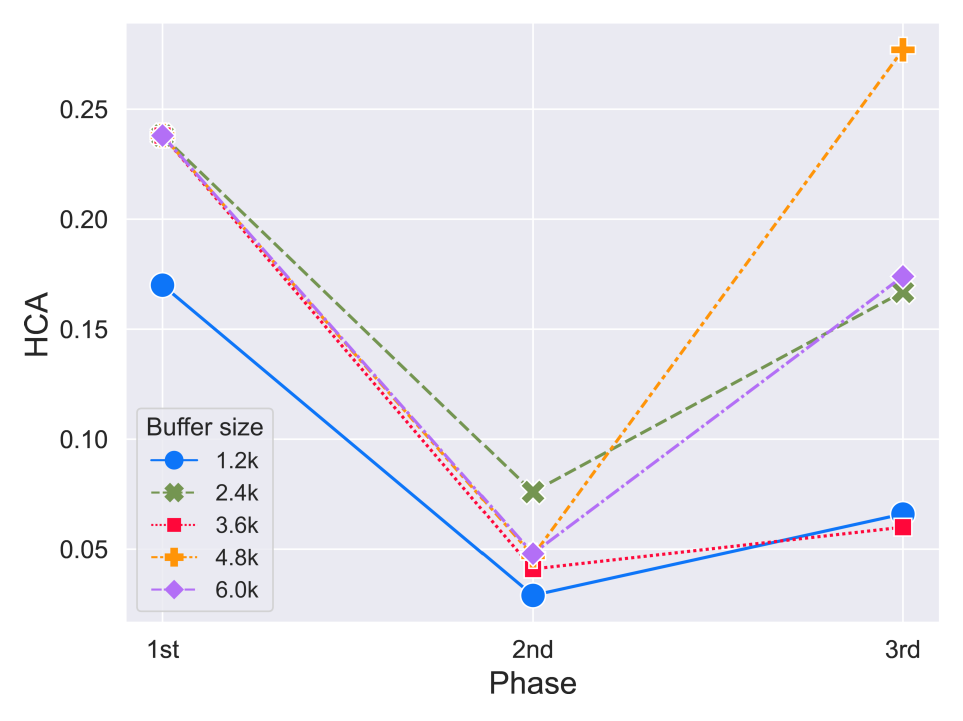}}
		\caption{Performance of $\text{OpenGCD}_\text{XGB}$ for IL, OSR, and GCD on CIFAR10 (top), CIFAR100 (center), and CUB (bottom) with different $|\mathcal{M}_r|$. Here, Acc is the average accuracy of all available test sets.}
		\label{fig:M}
	\end{figure}

	\vspace{-3mm}
	\paragraph{Parameter $\bm{|\mathcal{M}_r|}$}
	In the main paper, we report results at $|\mathcal{M}_r|\!=\!N_0$ on each dataset (20k$/$20k$/$2.4k for CIFAR10$/$CIFAR100$/$CUB). The effect of different $|\mathcal{M}_r|$ is shown in Fig. \ref{fig:M}. It can be found that Acc and HNA show a continuous and fluctuating decreasing trend over time, respectively. This is not related to the dataset type and buffer size but due to the increasing difficulty of the task. In comparison, the change in HCA appears to be random, and the non-robust estimation of the number of classes due to data variation may be the main reason. Moreover, from the first two columns of subfigures, it can be seen that increasing $|\mathcal{M}_r|$ improves performance but at a decreasing rate. From the perspective of computational overhead, storage burden and overall performance, $|\mathcal{M}_r|\!=\!N_0$ is appropriate.
	
\begin{figure}[!tbp]
		\centering
		\subfigure[]{\includegraphics[scale=0.28]{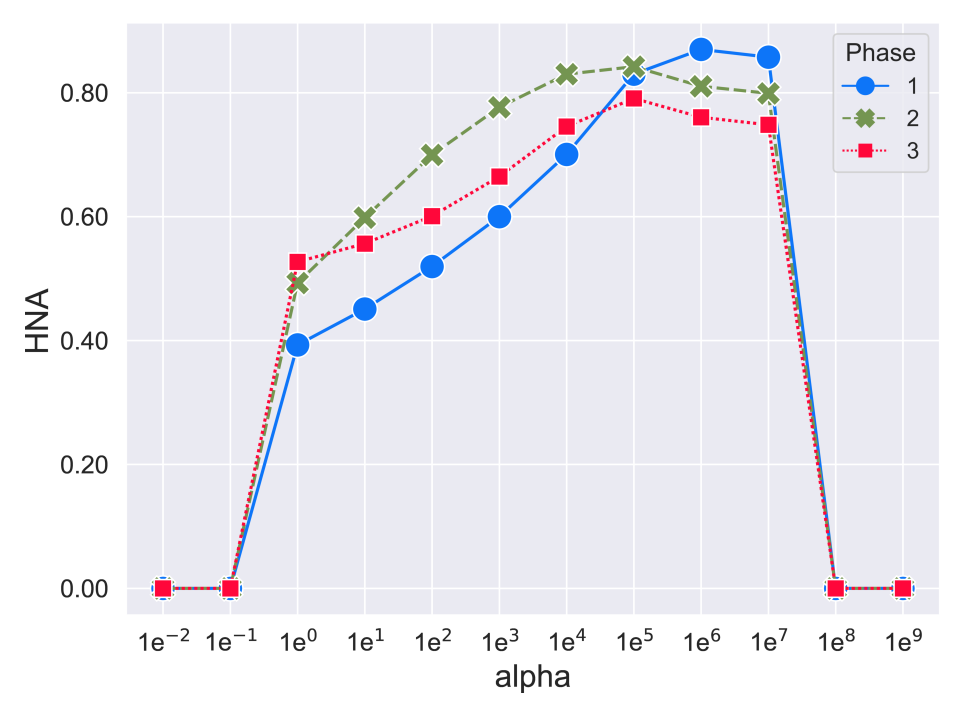}}
		\subfigure[]{\includegraphics[scale=0.28]{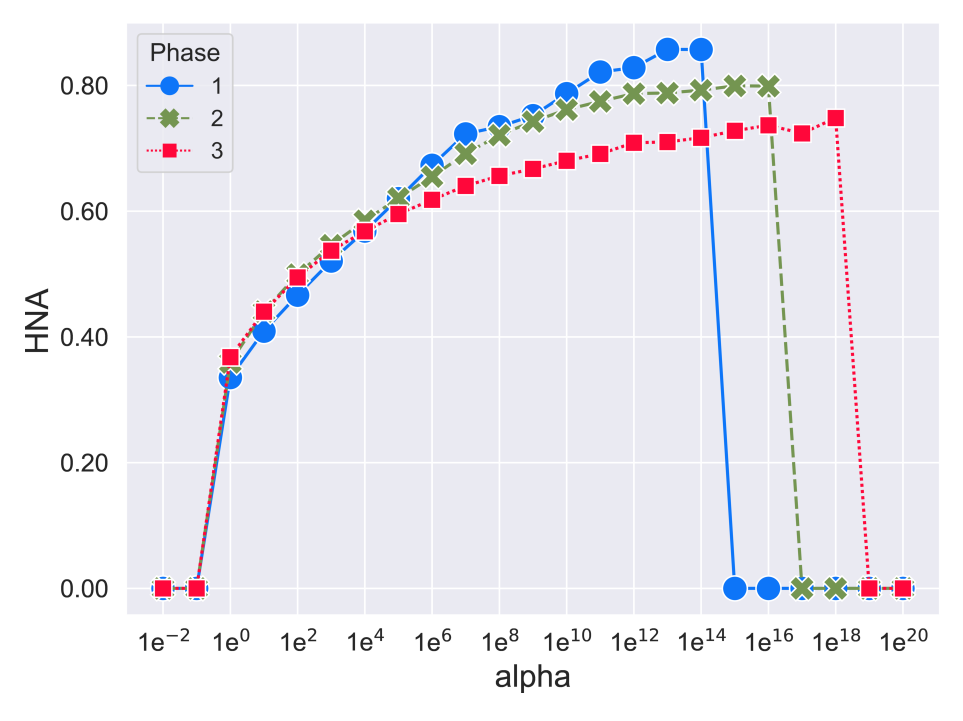}}
		\subfigure[]{\includegraphics[scale=0.28]{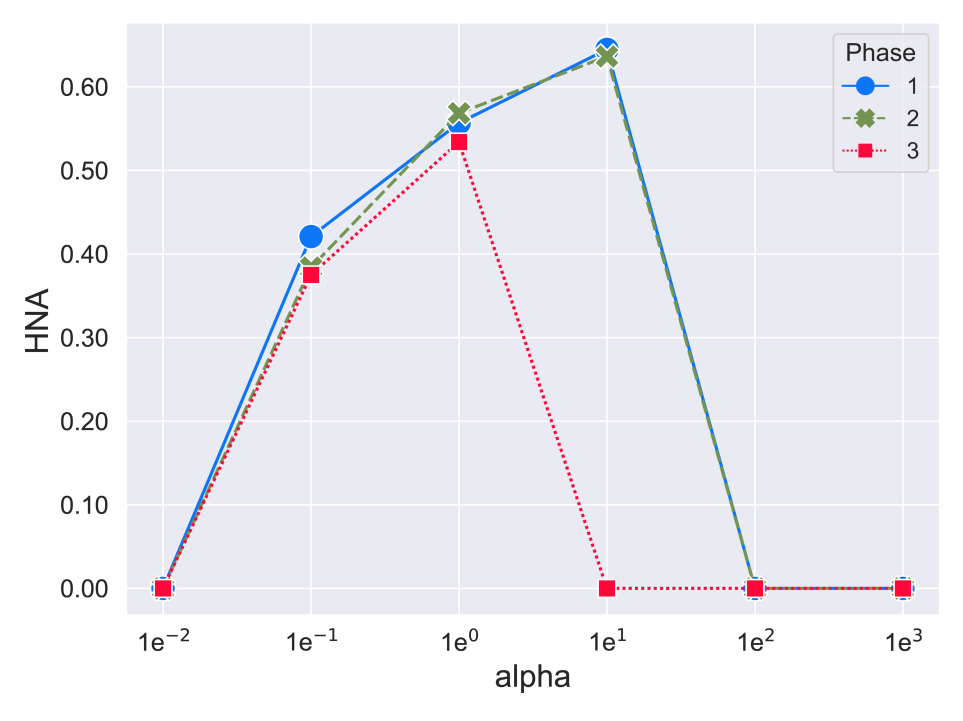}}
		\caption{Performance of $\text{OpenGCD}_\text{XGB}$ for	OSR on CIFAR10 (left), CIFAR100 (center), and CUB (right) with different $\alpha$.}
		\label{fig:alpha}
	\end{figure}

	\vspace{-3mm}
	\paragraph{Parameter $\bm{\alpha}$}
	In the main paper, we report the results on each dataset using $\alpha$ determined by the open set grid search protocol ($\{1e^5,1e^5,1e^5\}/\{1e^6,1e^6,1e^6\}/\{1,1,1\}$ for CIFAR10$/$CIFAR100$/$CUB). The effect of different $\alpha$ is shown in Fig. \ref{fig:alpha}. It can be found that with the increase of $\alpha$, HNA presents a trend of rising first and then falling. The $\alpha$ determined on CIFAR10 and CUB are close to or even equal to the optimal point, but is poor on CIFAR100. However, for CIFAR100, the performance gain from adding $\alpha$ is not significant and may also cause an inflated false alarm rate. Overall, the open set grid search protocol is still recommended.
	
	\section{Limitations and potential negative societal impacts}\label{sec:D}
	Although our method offers excellent compatibility and achieves state-of-the-art performance on public datasets, the performance still notably lags behind that of fully supervised models, especially when there are more classes. Moreover, since our method is developed towards the open world, real-world data is much more complex than the curated data used in our experiments, and periodic updates and performance degradation may also be unacceptable. Therefore, our method is not expected to provide sufficiently reliable inferences in safety-critical situations, such as autonomous driving and medical image analysis. Thus, careful validation should be performed on specific application scenarios before deployment to any real-world environment. Moreover, our method fails in cases where we cannot record or request data for novel classes encountered during online.
	
\end{document}